\documentclass[10pt,twocolumn,letterpaper]{article}

\usepackage[pagenumbers]{cvpr}
\usepackage{fancyhdr}
\usepackage[T1]{fontenc}
\usepackage{booktabs}
\usepackage{tabularx}
\usepackage{array}
\usepackage{fvextra}
\usepackage{enumitem}
\usepackage{listings}
\usepackage[most]{tcolorbox}

\newtcblisting{promptbox}[1]{
    enhanced,
    breakable,
    listing only,
    listing engine=listings,
    colback=black!2,
    colframe=black!35,
    colbacktitle=black!9,
    coltitle=black,
    boxrule=0.5pt,
    arc=1mm,
    left=2mm,
    right=2mm,
    top=1.5mm,
    bottom=1.5mm,
    before skip=6pt,
    after skip=6pt,
    fonttitle=\bfseries\footnotesize,
    title={#1},
    listing options={
        basicstyle=\ttfamily\scriptsize,
        breaklines=true,
        columns=fullflexible,
        keepspaces=true,
        showstringspaces=false
    }
}

\usepackage{multirow}

\usepackage{multirow}

\usepackage{caption}
\definecolor{cvprblue}{rgb}{0.21,0.49,0.74}
\usepackage[pagebackref,breaklinks,colorlinks,allcolors=cvprblue]{hyperref}

\def\confName{CVPR}
\def\confYear{2027}

\title{FigmaEval: A Benchmark for Evaluating AI Agents in Professional UI Design}

\author{%
\textbf{Xinyue Huang}\textsuperscript{1,2,*}\quad
\textbf{Yongyong Gao}\textsuperscript{1,3,*}\quad
\textbf{Zhengyuan Pan}\textsuperscript{1,*,\ensuremath{\dagger}}\\[0.2em]
\textbf{Zheng Wei}\textsuperscript{1}\quad
\textbf{Xinya Liu}\textsuperscript{1,5}\quad
\textbf{Duanyi Yan}\textsuperscript{1,4}\quad
\textbf{Zhiyu Hang}\textsuperscript{1,6}\\[0.2em]
\textbf{Zheng Ma}\textsuperscript{1,\ensuremath{\ddagger}}\quad
\textbf{Lewei Lu}\textsuperscript{1,\ensuremath{\ddagger}}\\[0.45em]
{\normalsize \textsuperscript{1}SenseTime\quad
\textsuperscript{2}Sun Yat-sen University}\\
{\normalsize \textsuperscript{3}Southern University of Science and Technology\quad
\textsuperscript{4}Tsinghua University}\\
{\normalsize \textsuperscript{5}East China University of Science and Technology\quad
\textsuperscript{6}University of Edinburgh}%
}
\hypersetup{pdfauthor={Xinyue Huang, Yongyong Gao, Zhengyuan Pan, Zheng Wei, Xinya Liu, Duanyi Yan, Zhiyu Hang, Zheng Ma, Lewei Lu}}
\makeatletter
\patchcmd{\@maketitle}{\begin{center}}{%
  \noindent\includegraphics[width=2.8cm]{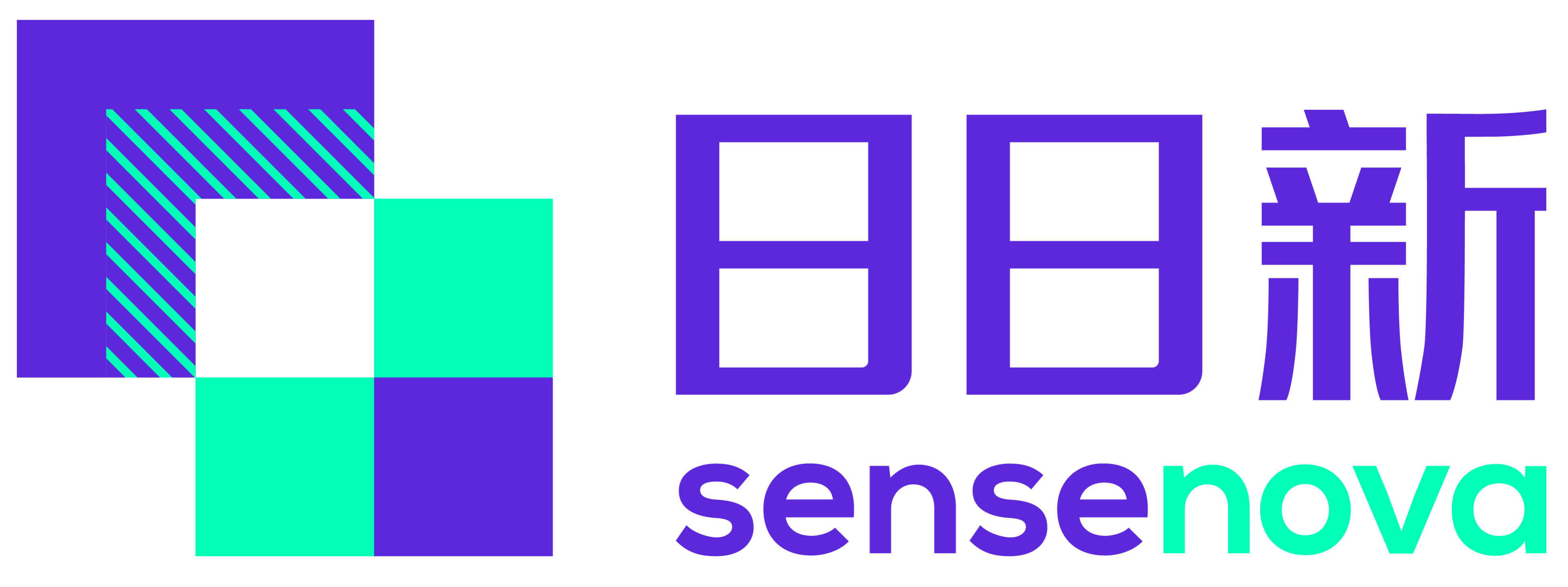}\par
  \vspace{0.6em}\begin{center}%
}{}{\PackageError{figmaeval}{Unable to place title logo}{}}
\makeatother

\begin{document}
\maketitle
\thispagestyle{fancy}

\begingroup
\renewcommand{\thefootnote}{}
\footnotetext{\textsuperscript{*}Equal contribution\quad \textsuperscript{\ensuremath{\dagger}}Project lead\quad \textsuperscript{\ensuremath{\ddagger}}Corresponding authors.}
\footnotetext{Our benchmark, code, model, and dataset are available at \url{https://github.com/PCNNgyy/figmaeval}.}
\endgroup


\begin{abstract}
Recent advances in AI agents have enabled autonomous UI design using professional tools such as Figma, yet evaluating their design capabilities remains challenging due to the open-ended nature of design tasks and the absence of unique ground truths. We introduce \textbf{FigmaEval}, a benchmark for evaluating AI agents on professional UI design tasks, comprising 84 tasks and 783 screens derived from real-world Figma design files. FigmaEval adopts a constrained-specification, open-solution paradigm, standardizing task requirements and design resources while allowing agents to independently determine layouts, visual hierarchies, component composition, and visual presentation. To assess design quality beyond reference similarity, we develop 22 expert-grounded evaluation rubrics across six dimensions and introduce a human evaluation arena where professional designers compare agent-generated designs through blinded pairwise judgments. Evaluating eleven models under a unified Figma agent setup, we find substantial agreement between rubric-based assessments and human preferences, alongside distinct differences in design practices and quality--efficiency trade-offs. FigmaEval provides a framework for systematically assessing professional UI design capabilities and understanding the strengths and limitations of current AI agents.
\end{abstract}

\section{Introduction}
\label{sec:introduction}

\begin{figure*}[t]
  \centering
  \includegraphics[width=\textwidth]{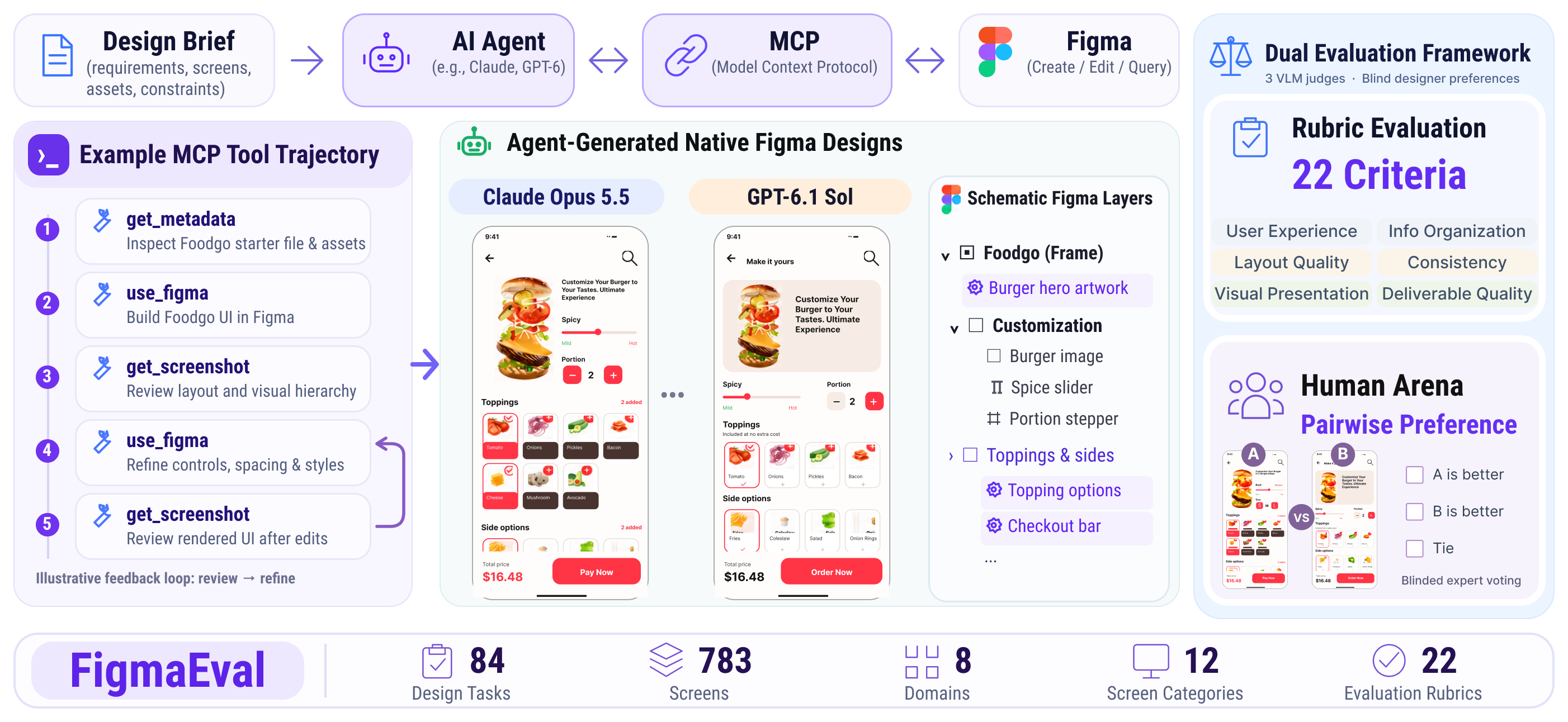}
  \caption{\textbf{Overview of FigmaEval.} Given a standardized design brief, AI agents interact with Figma through MCP to iteratively inspect, create, and refine native design artifacts. FigmaEval evaluates both rendered interface quality and underlying Figma structures using 22 expert-grounded rubrics, complemented by blinded pairwise comparisons from professional designers. The MCP interaction sequence and Figma layer tree are schematic illustrations.}
  \label{fig:teaser}
\end{figure*}

User Interface (UI) design builds on iterative design practices in which sketches and prototypes support the exploration and refinement of alternatives~\cite{goldschmidt1991,ferreira2007,buxton2010}. Research on creativity-support tools and professional design practice examines how tools support exploration and how designers apply design knowledge~\cite{resnick2005,stolterman2012,son2024}. In March 2026, Figma enabled AI agents to create and modify native design artifacts directly through its official Model Context Protocol (MCP) server~\cite{figma2026agents}. This development raises an important question: \emph{how can we evaluate agents' ability to autonomously produce professional-quality UI designs in Figma?}

Existing benchmarks do not fully address this question. Code-oriented benchmarks primarily evaluate frontend implementation, such as HTML and CSS generation~\cite{si2025design2code,wang2026figmabench}. Design Arena~\cite{designarena} evaluates human preferences across generated websites, applications, and presentations, but does not specifically assess native, editable Figma design artifacts. CANVAS~\cite{jeong2026canvas} evaluates tool-based UI design through replication and modification tasks, where performance is measured against predefined reference designs or target states. However, professional UI design often requires translating product requirements into original design solutions, rather than reproducing existing interfaces. Evaluating agents' autonomous design capabilities in Figma therefore remains underexplored.

Evaluating this capability presents two key challenges. First, unconstrained design briefs allow agents to vary the scope and complexity of their solutions, undermining fair comparison, while fixed reference designs restrict design freedom. Second, the same requirements may admit multiple valid design solutions, making reference-based similarity metrics insufficient for assessing professional design quality. Moreover, Figma design quality encompasses not only visual appearance but also the structural organization, reusability, and editability of design artifacts.

To close this gap, we present \textbf{FigmaEval}, a benchmark for autonomous, open-ended UI design in Figma. As illustrated in Fig.~\ref{fig:teaser}, FigmaEval provides a unified benchmarking framework in which agents iteratively create native, editable Figma artifacts from standardized design briefs, with their outputs evaluated through complementary rubric-based and human-preference assessments. We adopt a \emph{constrained-specification, open-solution} paradigm that standardizes product requirements, functionality, content, and design resources while preserving agents' freedom in design decisions. FigmaEval comprises \textbf{84 tasks and 783 screens} across eight application domains and twelve screen categories. To evaluate diverse design solutions without relying on reference layouts, we develop \textbf{22 expert-grounded rubrics} assessing interface usability, visual quality, and design deliverable quality through both rendered interfaces and underlying Figma structures. We further establish \textbf{Figma Design Arena}, an independent human-preference evaluation in which professional designers conduct blinded pairwise comparisons of agent-generated Figma artifacts, complementing fine-grained rubric-based assessment.

We evaluate eleven models under a unified Figma agent setup. Our analysis reveals two key findings: (i) high-quality Figma designs are accompanied by more systematic design practices, with the most preferred model consistently inspecting design systems and leveraging auto layout, reusable components, and shared styles; and (ii) greater generation time or API cost does not necessarily yield higher design quality, revealing distinct quality--efficiency trade-offs. Furthermore, rubric-based evaluations show strong alignment with professional designer preferences, achieving Kendall's $\tau_b = 0.764$ at the model-ranking level, and $76.58\%$ agreement on individual design comparisons. Failure analysis further identifies recurring limitations in asset adaptation, data-to-visual consistency, and complex content organization.The main contributions are summarized as follows:
\begin{itemize}
    \item 
We introduce \textbf{FigmaEval}, the first benchmark for evaluating AI agents' autonomous, open-ended creation of native, editable UI designs in Figma, balancing design freedom with task comparability.

    \item 
    We develop 22 fine-grained evaluation rubrics and establish an independent Figma Design Arena, providing complementary rubric-based assessments and expert preference rankings with strong empirical agreement.

    \item 
    We evaluate eleven models and investigate their design capabilities, quality--efficiency trade-offs, design behaviors, and representative failure modes, revealing strengths and limitations beyond aggregate rankings.
\end{itemize}

\section{Related Work}

\subsection{UI Generation Benchmarks.}
Early work such as pix2code~\cite{beltramelli2018pix2code} explored generating interface code from screenshots. Design2Code~\cite{si2025design2code} benchmarks screenshot-to-code generation, while Sketch2Code~\cite{li2025sketch2code} evaluates sketch-to-webpage prototyping, including iterative refinement through simulated user feedback or clarification. Figma2Code~\cite{gui2026figma2code} and FigmaBench~\cite{wang2026figmabench} assess design-to-code generation from structured Figma inputs. Beyond code-based UI implementation, ReDesign~\cite{yun2026redesign} reconstructs editable design structures from raster images, while CANVAS~\cite{jeong2026canvas} evaluates tool-based UI design through reference-guided replication and modification tasks. In contrast, FigmaEval evaluates agents' ability to autonomously create native, editable Figma designs from standardized product requirements without prescribed reference layouts.

\subsection{UI Design Quality and Creativity Assessment.}
Beyond visual fidelity, research on web aesthetics distinguishes classical and expressive aesthetic dimensions~\cite{lavie2004aesthetics}. DesignScape~\cite{odonovan2015designscape} supports graphic design through interactive layout refinement and brainstorming suggestions. For UI assessment, UIClip~\cite{wu2024uiclip} predicts design quality and semantic relevance. Duan et al.~\cite{duan2024feedback} use GPT-4 to generate heuristic-based feedback on Figma mockups, while UICrit~\cite{duan2024uicrit} provides professional designers' critiques and quality ratings to improve automated UI feedback. AUV-Bench~\cite{deng2026auvbench} examines aesthetic scoring, diagnosis, and repair, whereas Design Creativity Bench~\cite{rusia2026designcreativitybench} evaluates originality, creative diversity, and requirement satisfaction.
Building on these efforts, FigmaEval extends professional UI design quality assessment to native, editable Figma artifacts, introducing expert-grounded rubrics that evaluate both visual quality and underlying design structure.

\subsection{Human Preference-Based Design Evaluation.}
Human preferences provide a complementary approach to evaluating open-ended outputs, as exemplified by Chatbot Arena's pairwise evaluation of language models~\cite{chiang2024chatbotarena}. In design, UI-Bench~\cite{jung2025uibench} collects blinded expert pairwise comparisons of generated websites, while Design Arena~\cite{designarena} evaluates a broad range of generated artifacts, including websites, mobile apps, presentations, and 3D assets, through pairwise preferences, aggregating them into Bradley--Terry-based Elo ratings for model ranking. FigmaEval incorporates blinded pairwise comparisons by professional designers to assess agent-generated Figma artifacts and independently validate its fine-grained rubric-based evaluation.

\section{Benchmark}
\label{sec:benchmark}

\begin{figure*}[t]
  \centering
  \includegraphics[width=1.0\textwidth]{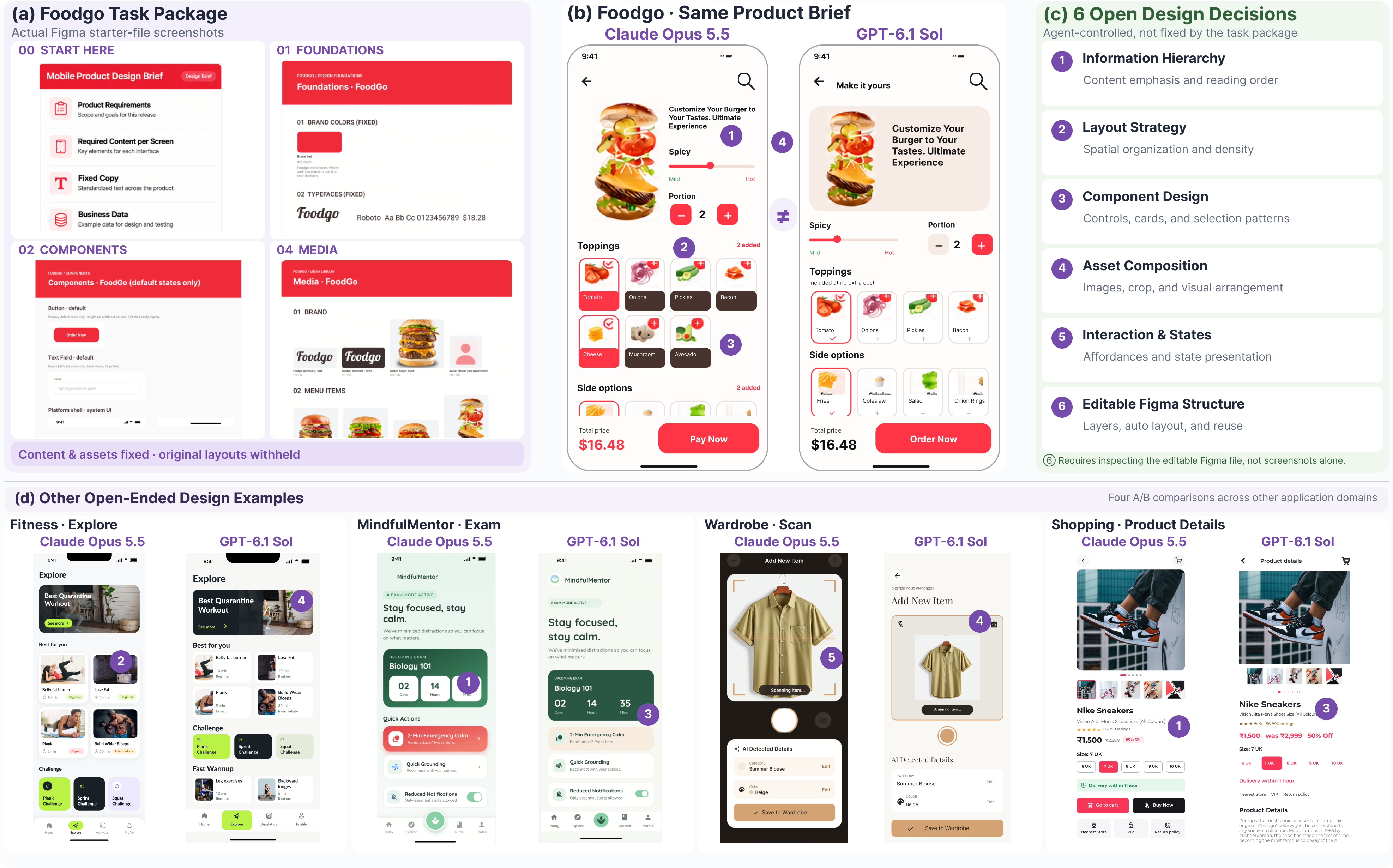}
  \caption{\textbf{Open-ended UI design under constrained specifications.} (a) The Foodgo task package provides fixed requirements, content, and design resources while withholding reference layouts. (b) Claude Opus 5.5 and GPT-6.1 Sol produce distinct design solutions from the same specification. (c) Six aspects of agent-controlled design decisions and native artifact structure. (d) Additional examples illustrate design variability across different application domains.}
  \label{fig:main-bench}
\end{figure*}

We introduce \textbf{FigmaEval}, a benchmark for evaluating AI agents' ability to autonomously create professional-quality, native, and editable UI designs in Figma. Rather than reproducing reference interfaces, FigmaEval requires agents to translate standardized product requirements into design solutions. The benchmark preserves design freedom while ensuring task comparability and capturing the complexity of real-world, multi-screen UI design.

\subsection{Task Formulation}
\label{sec:task_formulation}


Each task provides agents with a standardized product specification defining the required screens, functional requirements, and available design resources. Agents use Figma MCP to produce all required screens as native, editable Figma artifacts. The expected output is a coherent design deliverable rather than an executable application. Original reference layouts are withheld, requiring agents to independently determine how the specified requirements should be organized and presented. As illustrated in Fig.~\ref{fig:main-bench}, the same Foodgo task specification and shared design resources lead to visibly different design solutions. The agents make distinct decisions regarding information hierarchy, spatial organization, component composition, and visual asset placement, despite receiving identical design constraints. This example highlights the distinction between fixed task requirements and agent-controlled design decisions, which is central to FigmaEval's open-ended evaluation setting.

\subsection{Task Construction}
\label{sec:task_construction}

We develop a structured pipeline to transform existing UI designs into standardized, design-agnostic task specifications. The pipeline comprises four stages: source curation and scope selection, requirement abstraction and data normalization, design resource preparation, and quality validation.

\noindent\textbf{Source Curation and Scope Selection.}
We curate 84 public design files from Figma Community, prioritizing coherent product scenarios with multiple high-fidelity screens, meaningful functional workflows, and sufficient design complexity. We exclude isolated single-screen concepts, wireframes, component libraries, and incomplete or redundant designs. For each source file, we establish a consistent task scope by identifying the target platform, visual theme, and required screens.

\noindent\textbf{Requirement Abstraction and Data Normalization.}
We employ AI-assisted extraction to identify product context, screen requirements, functional elements, navigation relationships, and interface content from the source designs. Professional designers then review and refine the extracted information into structured task specifications. Rather than prescribing the visual composition of the original interfaces, each screen is described in terms of its intended goal, mandatory content, and functional relationships, leaving their visual realization open.

\noindent\textbf{Design Resource Preparation.}
To reduce variation arising from external asset acquisition, we provide essential branding constraints and basic interface elements through dedicated Foundations and Components pages. Reference-specific visual elements, such as decorative containers, state indicators, and embedded layout compositions, are excluded to avoid implicitly prescribing the original design solution.

\noindent\textbf{Quality Validation.}
Professional designers review the resulting specifications against the source designs to verify requirement completeness, factual correctness, and cross-screen consistency. The validation process further examines whether agents can identify all required deliverables from the specifications and whether multiple visually distinct solutions remain feasible.

\noindent\textbf{Task Packaging.}
Following construction and validation, each task is packaged as a read-only Figma Starter File comprising four pages. \textbf{00 Start Here} contains the product brief, required screens, functional requirements, fixed content and data, and navigation relationships. \textbf{01 Foundations} provides essential branding constraints, including primary colors and typography. \textbf{02 Components} supplies basic interface elements. \textbf{04 Media} contains organized visual assets, including images, logos, illustrations, and icons.

\begin{figure}[t]
  \centering
  \includegraphics[width=\linewidth]{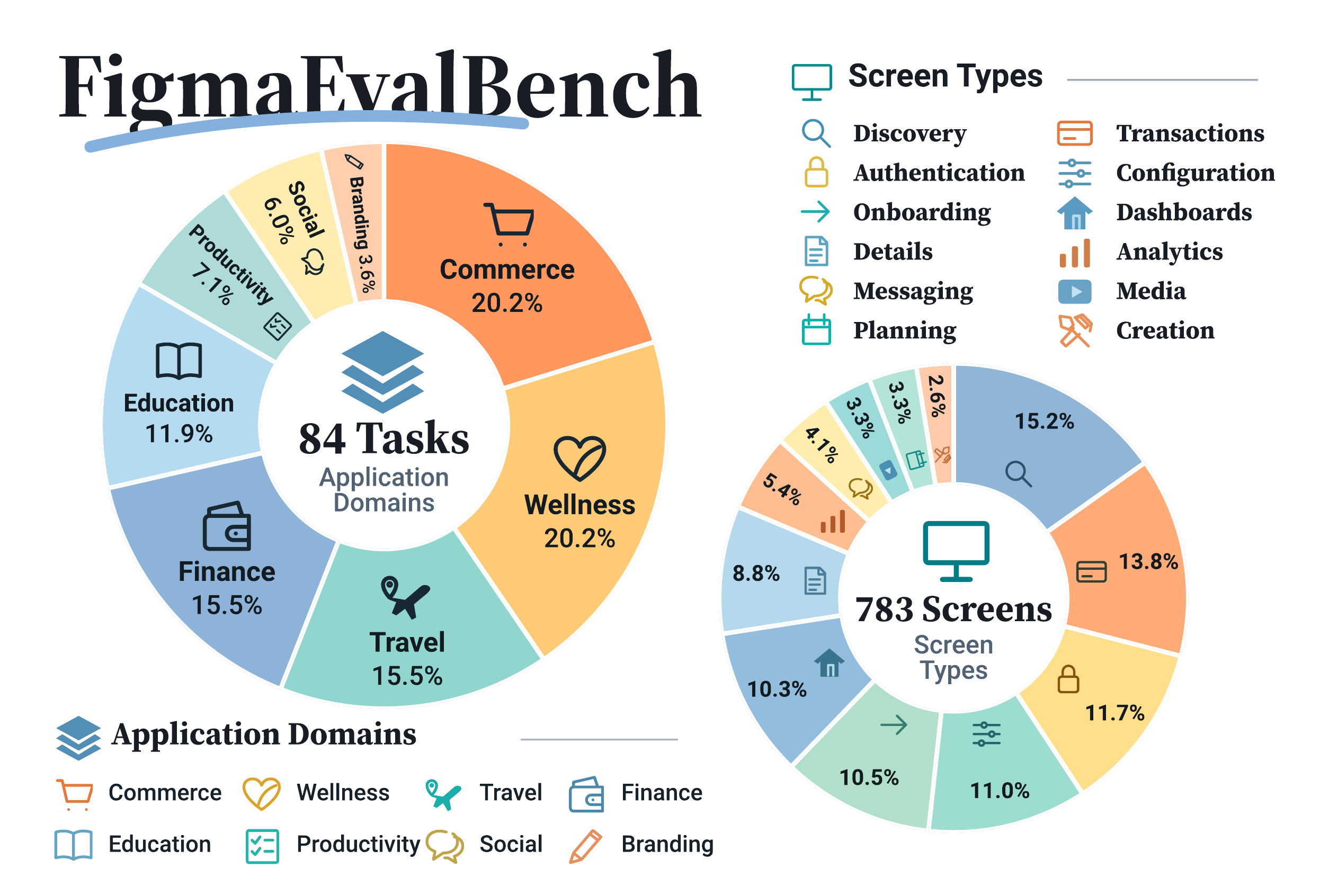}
\caption{\textbf{Composition of FigmaEval.} The benchmark comprises 84 design tasks across eight application domains (left) and 783 required screens across twelve screen categories (right). Percentages are computed over tasks and screens, respectively.}
  \label{fig:classic}
\end{figure}

\subsection{Benchmark Composition}
\label{sec:benchmark_composition}

FigmaEval comprises \textbf{84 UI design tasks requiring a total of 783 screens}, averaging 9.3 screens per task. As illustrated in Fig.~\ref{fig:classic}, the benchmark spans eight application domains and twelve screen categories, covering diverse product contexts and interface functionalities.

\begin{table}[t]
    \centering
    \caption{Overview of the 22 expert-grounded evaluation rubrics across six design quality dimensions. Full rubric definitions and scoring guidelines are provided in the supplementary material.}
    \label{tab:evaluation_rubrics}
    \scriptsize
\setlength{\tabcolsep}{2pt}
\renewcommand{\arraystretch}{1.05}
    \begin{tabularx}{\columnwidth}{
        @{}
        >{\raggedright\arraybackslash}p{0.25\columnwidth}
        >{\raggedright\arraybackslash}X
        >{\raggedleft\arraybackslash}p{0.18\columnwidth}
        @{}
    }
        \toprule
        \textbf{Dimension} &
        \textbf{Evaluation Focus} &
        \textbf{\shortstack[r]{Rubrics /\\Max Score}} \\
        \midrule

        User Experience &
        Discoverability of key content and actions;
        clarity of control affordances &
        4 / 8 \\
        \addlinespace[3pt]

        Information Organization &
        Logical grouping, reading hierarchy,
        semantic correctness, and action prioritization &
        5 / 15 \\
        \addlinespace[3pt]

        Layout Quality &
        Alignment, spacing, proportions, and whitespace
        utilization at component and page levels &
        4 / 12 \\
        \addlinespace[3pt]

        Consistency &
        Uniform component semantics and styling;
        consistent navigation and visual language
        across screens &
        2 / 4 \\
        \addlinespace[3pt]

        Visual Presentation &
        Typography, color roles, state semantics,
        and overall visual composition &
        5 / 10 \\
        \addlinespace[3pt]

        Deliverable Quality &
        Meaningful layer naming and structural hierarchy
        in editable Figma artifacts &
        2 / 4 \\

        \midrule
        \textbf{Total} & &
        \textbf{22 / 53} \\
        \bottomrule
    \end{tabularx}
\end{table}

\subsection{Design Evaluation Framework}
\label{sec:design_evaluation}

We adopt two complementary evaluation approaches:
rubric-based VLM assessment for fine-grained design
quality evaluation and independent human preference
evaluation for holistic design judgments.

\noindent\textbf{Rubric-Based VLM Evaluation.}
Drawing on Nielsen's usability heuristics~\cite{nielsen1994heuristics},
Atomic Design principles~\cite{frost2016atomic}, and recurring
design deficiencies identified by professional designers
through inspection of agent-generated Figma artifacts,
we develop 22 expert-grounded evaluation rubrics with
a maximum total score of 53 points.
These rubrics cover three aspects---interface usability,
visual quality, and design deliverable quality---organized
into six dimensions, as summarized in
Table~\ref{tab:evaluation_rubrics}.
Each rubric specifies an evaluation target, quality levels,
and observable evidence.
Criterion-based language-model evaluation has been studied
for natural language generation~\cite{liu2023geval}; here,
the criteria target visual and structural properties of UI artifacts.
We evaluate each agent-generated Figma design artifact
as a whole, considering both rendered interface screenshots
and underlying Figma structures.
Interface usability and visual quality are primarily
assessed through screenshots, while deliverable quality
additionally examines layer naming and structural hierarchy.
Full rubric definitions and scoring guidelines are
provided in the supplementary material.

\noindent\textbf{Figma Design Arena.}
Following Design Arena~\cite{designarena}, we establish
Figma Design Arena to collect professional designers'
preferences through blinded pairwise comparisons of
agent-generated Figma artifacts.
Reviewers inspect the standardized product specification,
aligned screenshots, and editable Figma files, selecting
the preferable overall design deliverable based on
requirement completeness, information organization,
visual quality, and usability.
Model identities and automated scores are concealed,
and left--right placement is randomized.
Direct preference votes are aggregated using the
Bradley--Terry model~\cite{bradley1952rank} to derive
human-preference rankings, providing an independent
basis for comparison with rubric-based evaluation.

\section{Experiments}
\label{sec:experiments}

\subsection{Setup}
\label{sec:experimental_setup}

\noindent\textbf{Evaluated Models.}
We evaluate eleven models: Claude Sonnet 5.5, Claude Opus 5.5,
DeepSeek V4 Flash, GPT-6.1 Sol, GPT-6 Astra, Qwen 3.8 Max,
Fable 5.1, GLM-5.3 Flash, SenseNova 6.8 Flash, Gemini 3.5 Flash,
and Kimi K3.

\noindent\textbf{Agent Setup.}
Tool-using language agents can interleave reasoning with
environment actions and observations~\cite{yao2023react}.
All models operate within the same Claude Code agent harness.
To isolate native Figma design capabilities from code-to-design
pipelines, we restrict the available MCP tools to
\texttt{whoami}, \texttt{get\_metadata},
\texttt{get\_screenshot}, and \texttt{use\_figma}.
This configuration supports structural inspection, visual
feedback, and direct manipulation of Figma elements through
the Plugin API, while excluding auxiliary code-based workflows.

\noindent\textbf{VLM-as-a-Judge.}
We evaluate the generated designs using three VLM judges:
Claude Opus 5.5, GPT-6.1 Sol, and DeepSeek V4.1 Flash.
Judging is repeated across three seeded passes
(42, 102, and 423).
Prior work identifies systematic biases in model-based judges~\cite{zheng2023judging}, motivating validation against independent human judgments. We measure model-level ranking agreement using Kendall's $\tau$ and comparison-level agreement using the proportion of pairwise preferences consistent with human judgments.

\noindent\textbf{Evaluation Metrics.}
We assess human--VLM alignment at two levels.
At the model level, Kendall's $\tau$ measures the agreement
between model rankings derived from human preference
win rates and mean rubric scores.
At the design-pair level, we measure the proportion of within-task comparisons in which rubric-based preferences agree with human judgments.

\begin{table*}[t]
    \centering
    \caption{
        Main results for eleven evaluated models on \textsc{FigmaEval}.
        Human preference results include Arena win rates and Elo ratings.
        Rubric scores are evaluated on 84 tasks and averaged across three
        VLM judges, each with three scoring passes, then normalized to
        percentages; higher is better.
    }
    \label{tab:main-results}
    \includegraphics[width=\textwidth]{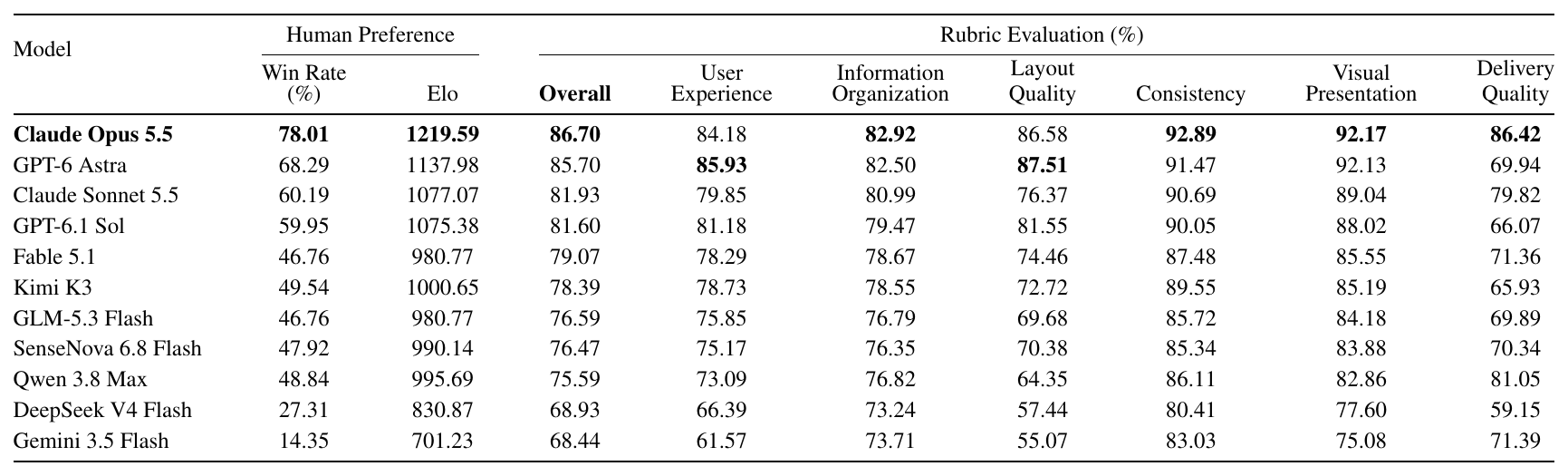}
\end{table*}

\subsection{Overall Evaluation}
\label{sec:benchmark_results}

\noindent\textbf{Overall Performance.}
Table~\ref{tab:main-results} shows that Claude Opus 5.5 and GPT-6 Astra form the leading group under both human preference and rubric-based evaluation, with Opus ranking first in both Elo (1219.59) and overall rubric score (86.70\%). However, aggregate rankings mask distinct capability profiles. GPT-6 Astra slightly surpasses Opus in user experience and layout quality, but scores lower than even Claude Sonnet 5.5 in deliverable quality. Across the six evaluation dimensions, layout quality exhibits the largest variation among models (55.07\%--87.51\%), suggesting that spatial organization remains a key differentiator in professional UI design.

\begin{figure}[t]
  \centering
  \includegraphics[width=\linewidth]{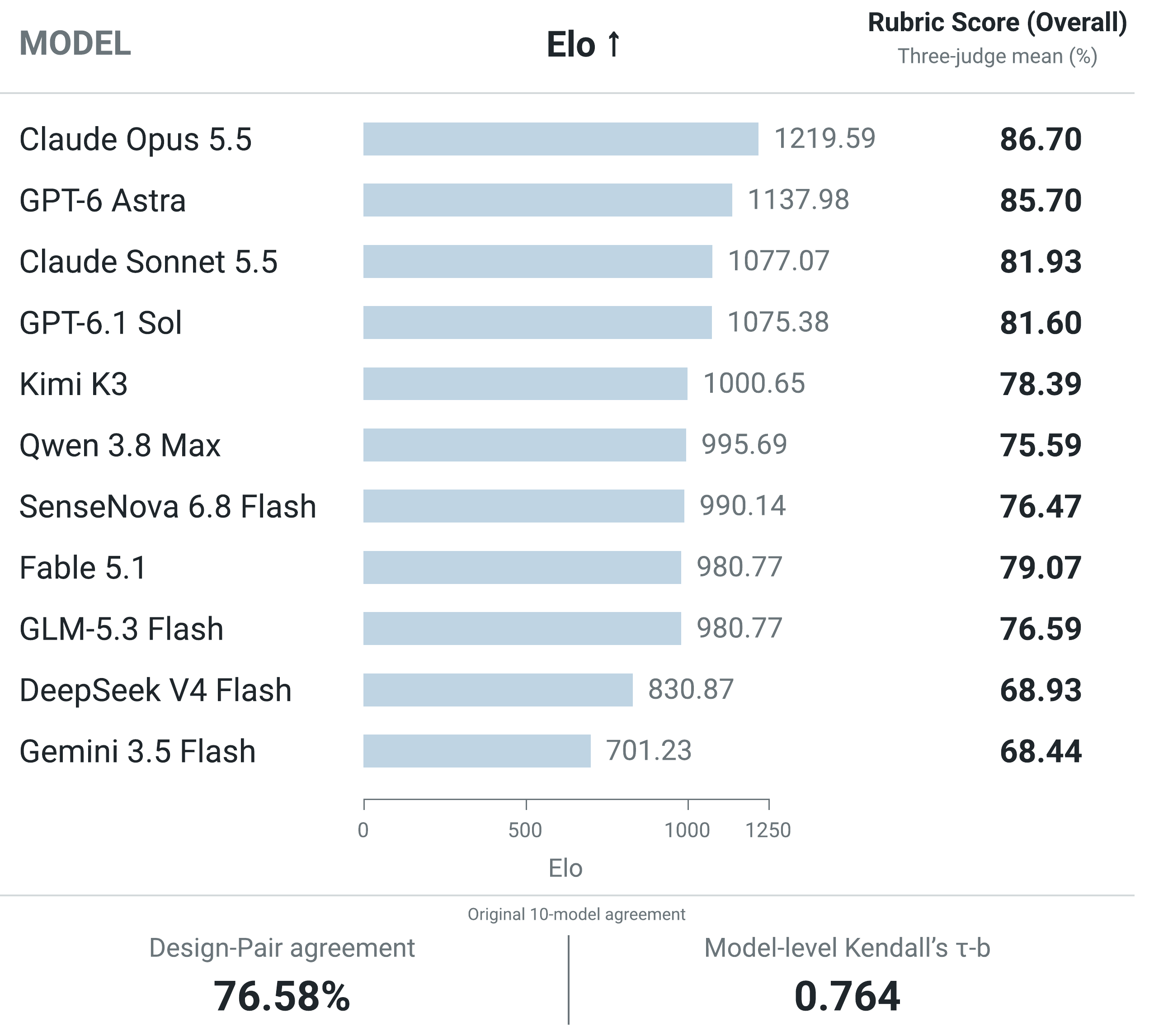}
\caption{Human preference and rubric-based evaluation of
eleven models. Agreement statistics refer to the original eleven-model subset.}
  \label{fig:alignment}
\end{figure}

\noindent\textbf{Alignment with Human Preferences.}
Fig.~\ref{fig:alignment} compares Elo-based human preference rankings with rubric scores averaged across three VLM judges. The two evaluations show substantial agreement at the model level (Kendall's $\tau_b = 0.764$), with Claude Opus 5.5 and GPT-6 Astra occupying the top two positions under both evaluations. At the design-pair level, rubric-based preferences agree with blinded professional judgments in 76.58\% of comparisons. Together, these results demonstrate substantial alignment between rubric-based evaluation and professional designer preferences at both the model-ranking and individual-comparison levels.

\begin{figure}[tbp]
  \centering
  \includegraphics[width=\columnwidth]{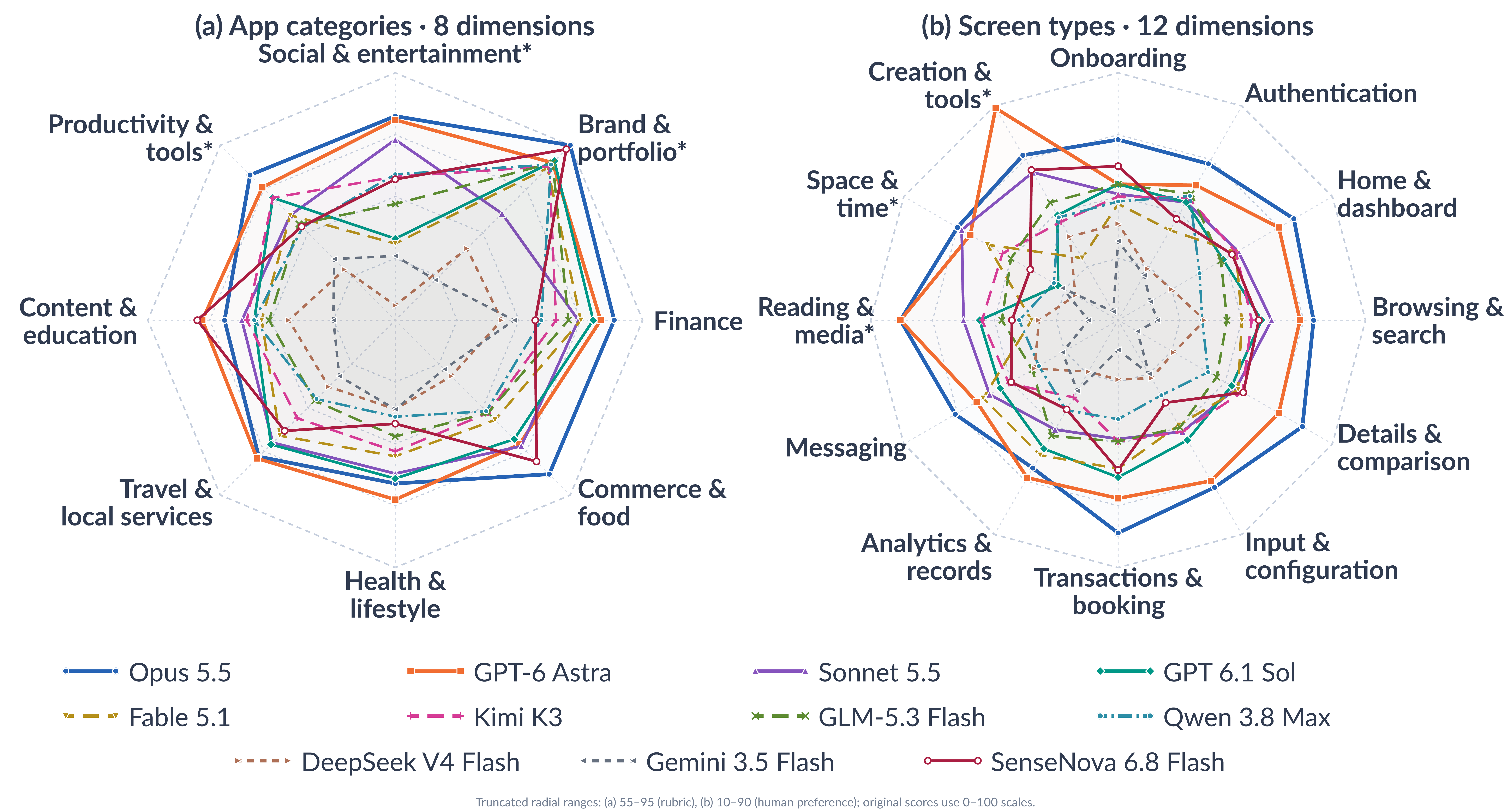}
  \caption{\textbf{Model performance across UI categories.}
Scores of eleven models: (a) rubric scores across eight application domains and (b) human preference scores across twelve screen types. Model performance follows broadly consistent trends across application domains, while larger differences emerge in screen types involving multiple interdependent design decisions.}
\label{fig:type}
\end{figure}

\noindent\textbf{Model Gaps Widen When Design Decisions Interact.}
Fig.~\ref{fig:type} shows that model performance remains broadly consistent across application domains, while differences become more pronounced across screen types that demand multiple design decisions simultaneously. Onboarding screens often center on a single message and primary action, whereas product details, transactions, and dashboards must balance information hierarchy, visual emphasis, spatial organization, and user decision flows. These requirements are interdependent: emphasizing one element can compromise another, and individually reasonable components can still form a confusing interface. Stronger agents distinguish themselves by resolving these competing demands into a coherent whole, rather than merely satisfying individual design requirements.

\begin{figure}[tbp]
  \centering
  \includegraphics[width=0.8\linewidth]{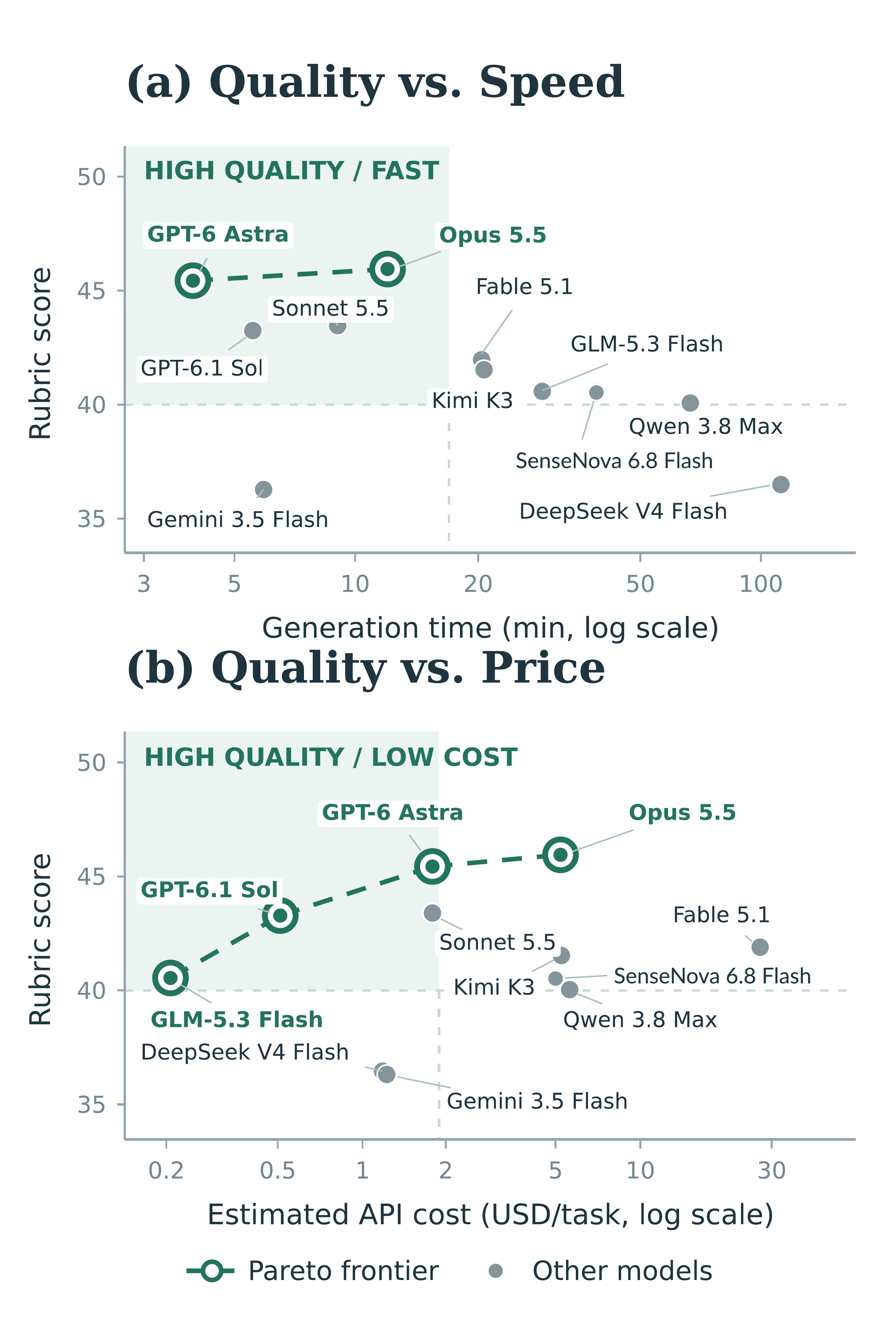}
\caption{\textbf{Quality--efficiency trade-offs.}
    Design quality versus (a) generation time
    and (b) estimated API cost across eleven models.
    Highlighted points indicate the Pareto frontier.}
  \label{fig:quality_efficiency}
\end{figure}

\subsection{Quality--Efficiency Trade-offs}

\textbf{Quality versus efficiency.}
Figure~\ref{fig:quality_efficiency} examines the relationship between design quality and resource expenditure, revealing that longer generation times or higher API costs do not necessarily yield better designs. For instance, DeepSeek V4 Flash requires substantially more time than several higher-performing models, while Fable 5.1 incurs greater API cost than Claude Opus 5.5 despite achieving lower design quality. More importantly, the Pareto frontiers reveal distinct quality--efficiency trade-offs. In terms of generation time (Fig.~\ref{fig:quality_efficiency}(a)), GPT-6 Astra approaches the rubric score of Claude Opus 5.5 (85.7\% vs.\ 86.7\%) with substantially lower latency, while Opus achieves the highest quality at greater time cost. Under API cost (Fig.~\ref{fig:quality_efficiency}(b)), GLM-5.3 Flash and GPT-6.1 Sol additionally emerge as Pareto-optimal alternatives, offering lower-cost operating points at different quality levels. These findings demonstrate that no single model is universally optimal; the preferred choice depends on the relative priorities of design quality, generation speed, and monetary cost.

\begin{figure*}[tp]
  \centering
  \includegraphics[width=0.85\textwidth]{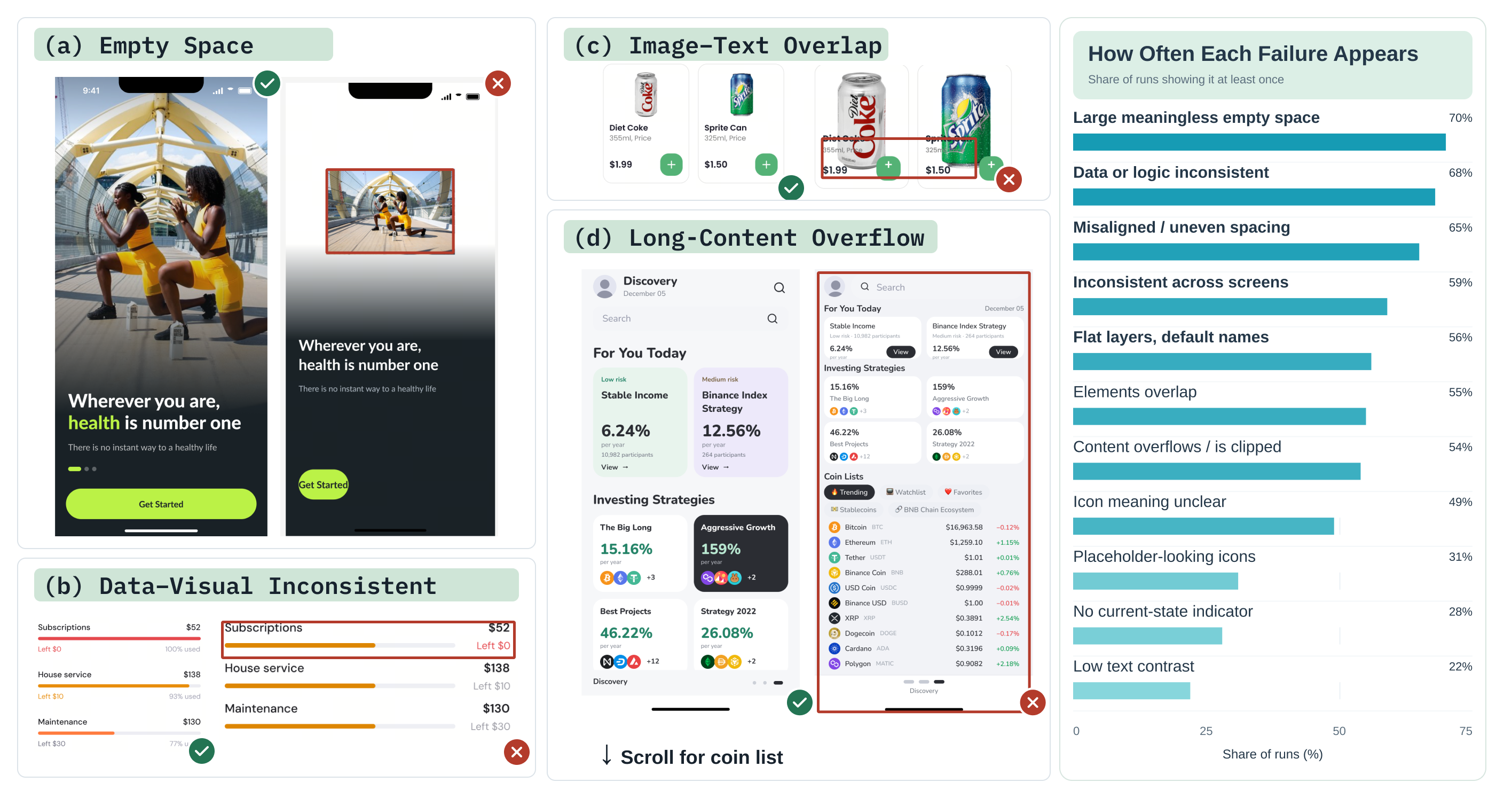}
\caption{\textbf{Representative failure modes of UI design agents.}
(a--d) Professional designers identify four representative failures through qualitative inspection of generated artifacts: excessive empty space caused by poor asset adaptation, inconsistent data-to-visual encoding, image--text overlap, and long-content overflow caused by inadequate content organization.  
(Right) Failure prevalence across evaluated runs, derived from the 22-criterion evaluation framework.}
\label{fig:failure_modes}
\end{figure*}

\subsection{Design Quality Diagnostics}
\label{sec:design_analysis}

\noindent\textbf{Design Practices of Leading Agents.}
To investigate the design practices associated with stronger performance, we analyze agents' Figma interaction trajectories, focusing on design-system inspection, layout construction, and component reuse. Before making modifications, Claude Opus 5.5 inspects the provided design system, including typography, variables, and reusable components, in 100\% of runs, compared with 14--29\% for GLM-5.3 Flash, Qwen 3.8 Max, and Fable 5.1. During construction, Opus performs an average of 0.95 auto-layout operations per newly created frame, names 6.6 layers per frame, and creates 20 components per screen. In contrast, both GPT models rely predominantly on absolute positioning, with minimal use of auto-layout. Opus also makes more systematic use of shared styles and variables across its designs.

These observations reveal a distinction between visual layout quality and the structural quality of editable design artifacts. Although GPT-6 Astra slightly outperforms Opus in layout quality (87.51\% vs. 86.58\%), Opus achieves substantially higher deliverable quality (86.42\% vs. 69.94\%) and ranks first in human preference. This contrast suggests that producing visually effective layouts and maintaining well-structured, reusable Figma artifacts represent related but distinct capabilities. Opus's design-system-oriented workflow is consistent with its stronger deliverable quality, highlighting the importance of evaluating not only rendered interfaces but also the underlying design structures.

\noindent\textbf{Representative Failure Modes.}
To investigate the limitations of current UI design agents, professional designers inspected generated artifacts and identified four representative failure patterns, as illustrated in Fig.~\ref{fig:failure_modes}. \textbf{(1) Empty Space:} agents fail to adapt provided images to target screen layouts, resulting in poorly scaled assets and excessive empty space. \textbf{(2) Image--text overlap:} visual elements obscure textual content, compromising readability. \textbf{(3) Data--visual Inconsistent:} visual representations contradict the underlying data, such as progress bars that fail to reflect actual values. \textbf{(4) Long-content Overflow:} agents compress extensive content into a single viewport rather than organizing it into scrollable regions. With additional examples provided in the supplementary material.

To examine whether these deficiencies extend beyond isolated cases, we further quantify failure prevalence across evaluated runs using our 22-criterion evaluation framework in Fig.~\ref{fig:failure_modes}. Excessive empty space (70\%), data or UI logic inconsistencies (68\%), and misalignment or uneven spacing (65\%) are among the most prevalent issues. Overlapping elements (55\%) and content overflow or clipping (54\%) also appear in more than half of the runs. These quantitative findings reinforce the qualitative observations, revealing recurring challenges in adapting visual assets, preserving data semantics, and organizing complex content. Together, they highlight the gap between generating visually plausible interfaces and delivering coherent, usable UI designs.

\newcommand{\rubricHolisticEnsemble}{65.6}
\newcommand{\rubricHolisticTau}{0.600}
\newcommand{\rubricSixEnsemble}{71.2}
\newcommand{\rubricSixTau}{0.733}
\newcommand{\rubricFullEnsemble}{76.58}
\newcommand{\rubricFullTau}{0.764}

\newcommand{\rubricvalue}[1]{#1}

\subsection{Ablation Study}
\label{sec:rubrics-ablation}
\paragraph{Rubric Granularity.} To assess the contribution of explicit evaluation criteria, we compare holistic scoring without predefined dimensions, scoring along the six design dimensions without detailed criteria or scoring anchors, and the full 22-criterion rubric. We hold the design artifacts, expert preferences, judges, and score aggregation fixed; the latter two schemes also share dimension weights. As shown in Table~\ref{tab:rubrics-ablation}, human agreement increases from \rubricHolisticEnsemble\% with holistic scoring to \rubricSixEnsemble\% with six-dimension scoring and \rubricFullEnsemble\% with the full rubric, while model-ranking agreement rises from $\tau_b=\rubricHolisticTau$ to $\rubricSixTau$ and $\rubricFullTau$, respectively. The full rubric improves human agreement by 10.98 percentage points over holistic scoring and 5.38 points over six-dimension scoring, suggesting that detailed criteria and scoring anchors provide useful guidance beyond broad design dimensions.

\begin{table}[t]
    \centering
    \caption{\textbf{Effect of evaluation rubrics on alignment with professional
    designers.} All schemes use the same design artifacts, overall expert
    preferences, and judge models. Scores are averaged over three passes
    per judge and then across the three judges.
    Higher is better for all metrics.
    Full-rubric ensemble values follow Fig.~\ref{fig:alignment}.}
    \label{tab:rubrics-ablation}
    \begingroup
    \small
    \setlength{\tabcolsep}{3pt}
    \renewcommand{\arraystretch}{1.12}
    \begin{tabular}{@{}lcc@{}}
        \toprule
        Evaluation scheme
        & \shortstack{Human agreement\\(\%)}
        & \shortstack{Kendall's\\$\tau_b$} \\
        \midrule
        Holistic scoring
        & \rubricvalue{\rubricHolisticEnsemble}
        & \rubricvalue{\rubricHolisticTau} \\
        Six-dimension scoring
        & \rubricvalue{\rubricSixEnsemble}
        & \rubricvalue{\rubricSixTau} \\
        Full rubric (22 criteria)
        & \rubricvalue{\rubricFullEnsemble}
        & \rubricvalue{\rubricFullTau} \\
        \bottomrule
    \end{tabular}
    \endgroup
\end{table}

\paragraph{Repair-Based Fine-Tuning. }We inject defects from 20 design-error types defined by our evaluation
rubric into Figma pages. Given only a page and its editing scope, with no
defect list, Claude-Opus~5, as the Teacher Model, locates and repairs the defects in place through
Figma MCP tool calls while preserving correct regions. This process yields
1{,}485 multimodal trajectories, comprising approximately 16{,}000 edit
calls and 5{,}880 screenshot calls, that record inspection, editing, and
visual verification. The training pages are disjoint from the evaluation
tasks. We use these trajectories for full-parameter supervised fine-tuning
of Qwen3.8-9B and compare the fine-tuned model with the base model under
the same evaluation setup. Table~\ref{tab:sft-preference} shows that, across 100 pairwise comparisons on held-out tasks, designers prefer the
fine-tuned model in 90.0\% of comparisons (95\% Wilson interval
[82.6, 94.5]) and the base model in 2.0\% ([0.6, 7.0]); the remaining
8.0\% ([4.1, 15.0]) are ties.

\begin{table}[t]

\centering
    \small
    \captionof{table}{Designer preferences for fine-tuned (SFT) and base
    Qwen3.8-9B over 100 pairwise comparisons on held-out tasks, where the
    30- and 50-comparison columns are illustrative draws without
    replacement from the observed outcome counts, restricted to SFT win
    rates above 87\%.}
    \label{tab:sft-preference}
    \begin{tabular}{@{}lccc@{}}
        \toprule
        Outcome (\%) & \shortstack{30}
                     & \shortstack{50}
                     & \shortstack{100} \\
        \midrule
        SFT wins  & 93.3 & 88.0 & \textbf{90.0} \\
        Tie       &  3.3 &  8.0 & 8.0 \\
        Base wins &  3.3 &  4.0 & 2.0 \\
        \bottomrule
    \end{tabular}
\end{table}\par

\section{Conclusion}
\label{sec:conclusion}

We introduced FigmaEval, a benchmark of 84 open-ended tasks for assessing the quality of editable UI designs produced by AI agents in Figma. Its 22 criteria assess interface usability, visual quality, and the quality of the design deliverable. The benchmark connects automated evaluation with professional designers' blind preferences and uses criterion-level feedback to guide supervised fine-tuning.


\clearpage
\section*{LLM Usage Statement}
\label{sec:llm-usage}

We used OpenAI Codex and Anthropic Claude for
research discussions, text and figure drafts, language
 polishing, experimental coding assistance, and visualization. All content was reviewed and verified
 by the authors, who take full responsibility for the
 work.

{
    \small
    \bibliographystyle{ieeenat_fullname}
    \bibliography{main}
}

\clearpage
\hypersetup{pageanchor=false}
\setcounter{page}{1}
\appendix
\twocolumn[\centering\Large\textbf{\thetitle}\\
\vspace{0.5em}Appendix\\
\vspace{1.0em}]

\section{Benchmark Construction and Agent Setup}
\label{sec:sup_benchmark}

\subsection{Task Construction and Validation}

FigmaEval starts from 84 public Figma Community files containing
multi-screen product workflows. For each file, we identify the target
platform and required screens, retain functionally distinct interaction
states, and consolidate redundant variants. AI-assisted extraction
produces an initial specification of screen goals, required content,
functional elements, and navigation relationships. Professional designers
review the specification against the source and correct omissions and
inconsistencies. Shared entities are normalized across screens so that,
for example, product quantities, prices, and transaction totals agree
throughout a workflow.

\begin{figure*}[!t]
  \centering
  \begin{minipage}[t]{0.57\textwidth}
    \vspace{0pt}
    \small
    \textbf{(a) 00 Start Here}\par
    \smallskip
    \textbf{Product brief.} Design the complete Foodgo mobile workflow,
    from browsing burgers and choosing portions and spice levels to
    customizing an order and completing payment.\par
    \medskip
    \begingroup
    \setlength{\fboxsep}{8pt}
    \fcolorbox{black!15}{black!4}{%
      \begin{minipage}{\dimexpr\linewidth-2\fboxsep-2\fboxrule\relax}
        \raggedright
        \textbf{Product detail (four screens)}\par
        \smallskip
        \textbf{Goal.} Evaluate a burger and choose the portion and spice
        level before ordering.\par
        \smallskip
        \textbf{Must Include.} Back, search, and Order Now actions.
        Show the burger image, full name, rating, preparation time, and
        description. Show the current portion with controls to increase
        or decrease it, the corresponding price, and the selected spice
        level on a Mild-to-Hot scale.\par
        \smallskip
        \textbf{Links.} Back returns to Home. Order Now opens Customize
        for the cheeseburger. On the other three screens, Order Now is
        an entry point only.
      \end{minipage}}
    \endgroup
  \end{minipage}\hfill
  \begin{minipage}[t]{0.39\textwidth}
    \vspace{0pt}
    \small
    \textbf{(b) 01 Foundations}\par
    \includegraphics[width=0.84\linewidth]{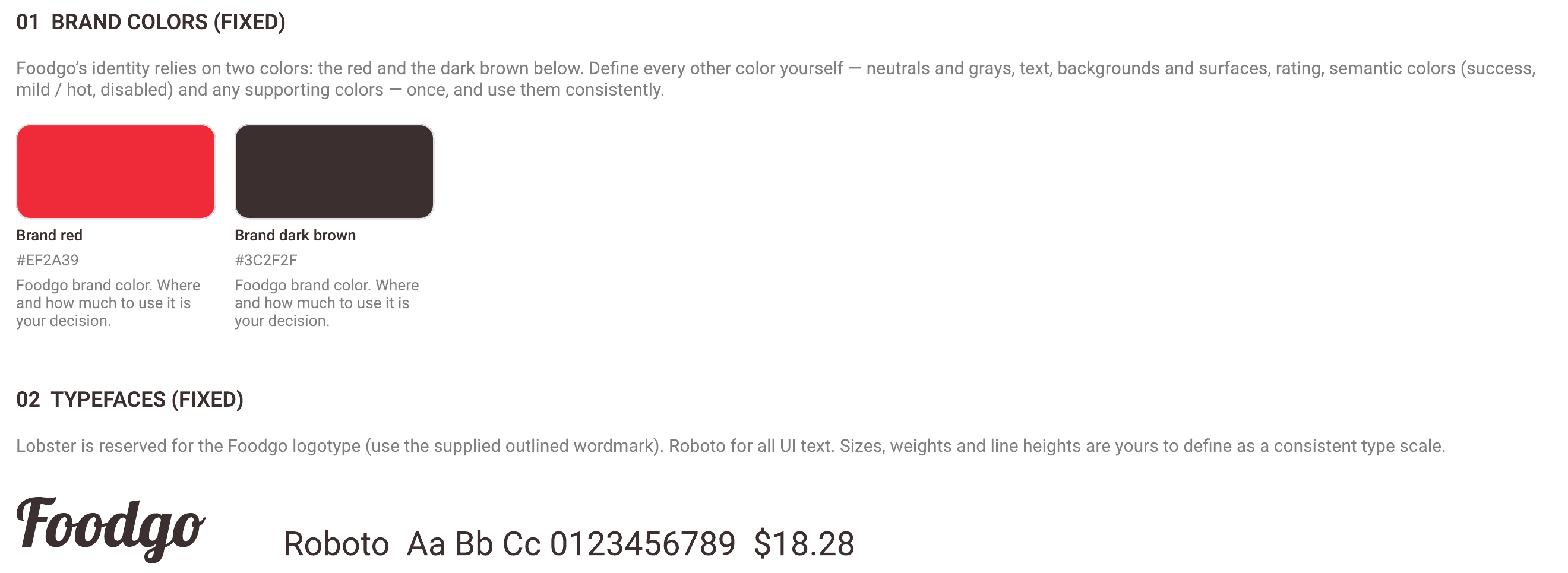}\par
    \smallskip
    \textbf{(c) 02 Components}\par
    \includegraphics[width=0.58\linewidth]{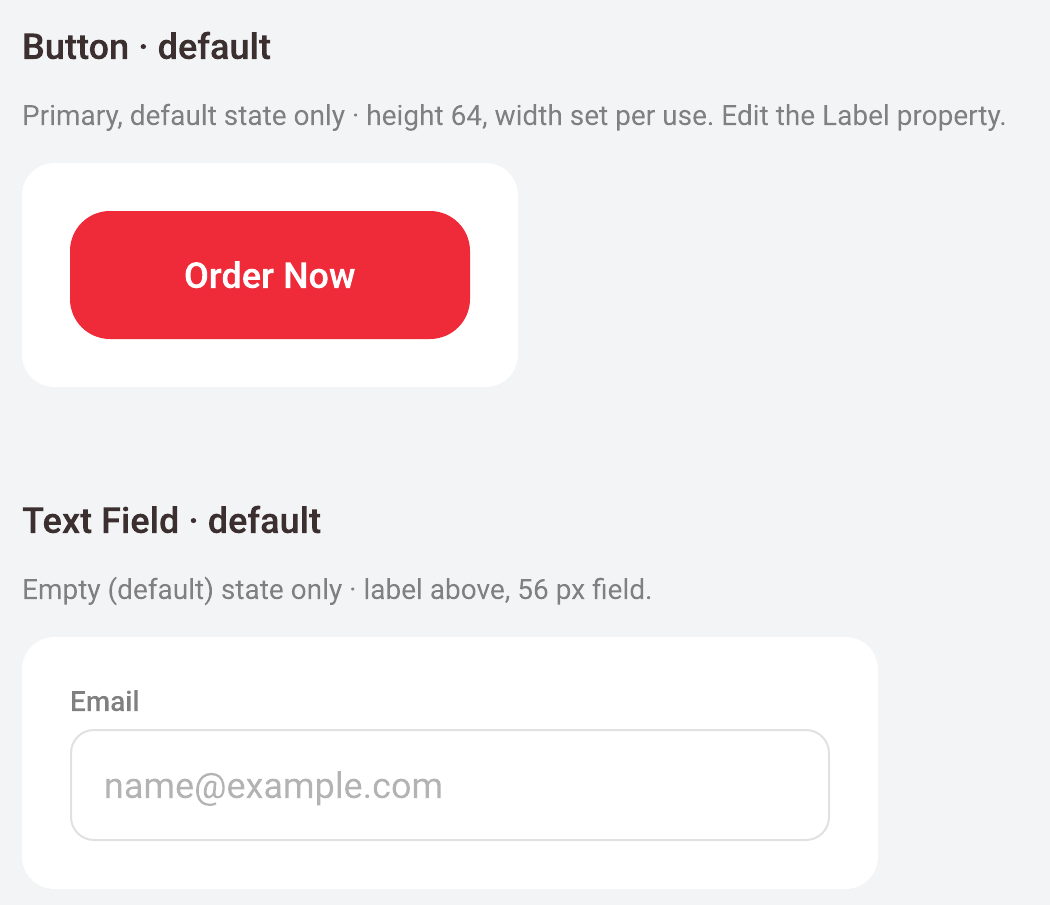}\par
    \smallskip
    \textbf{(d) 04 Media}\par
    \includegraphics[width=\linewidth]{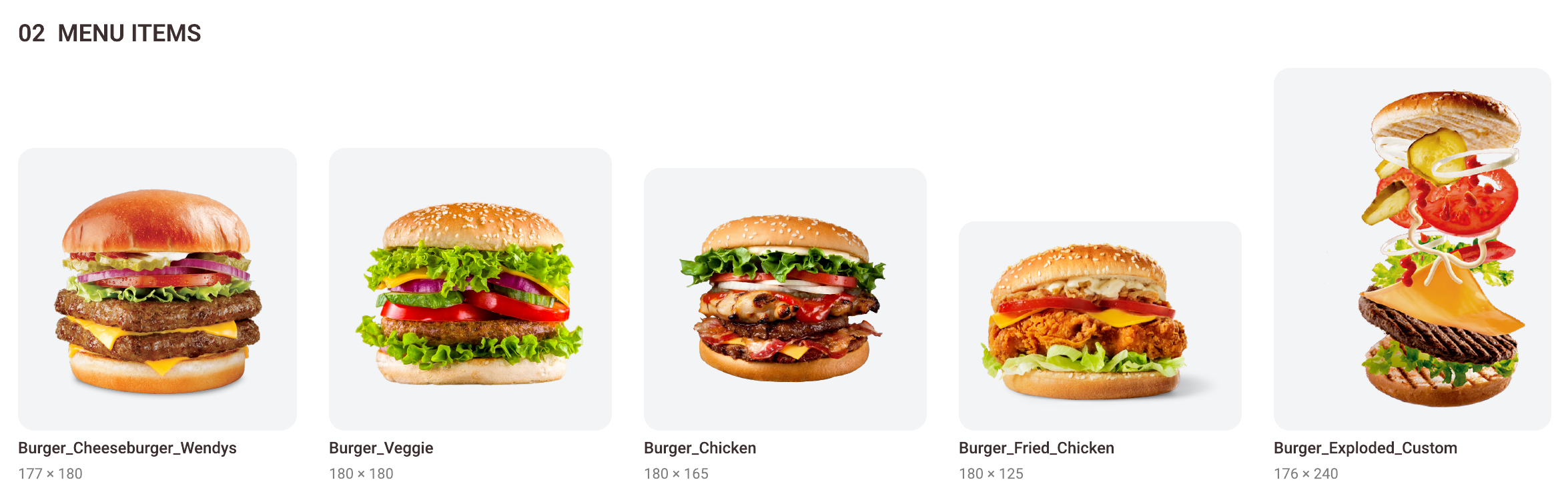}
  \end{minipage}
  \caption{\textbf{Foodgo starter file.} Panel (a) summarizes the
  product brief and product-detail task card. Panels (b)--(d) show
  excerpts of the supplied colors and typefaces, interface components,
  and menu images with asset identifiers.}
  \label{fig:sup_foodgo_starter}
\end{figure*}

Each read-only starter file contains four pages. \textbf{00 Start Here}
provides the product brief and per-screen cards with \emph{Goal},
\emph{Must Include}, and \emph{Links} fields. \textbf{01 Foundations}
provides brand colors and typography. \textbf{02 Components} supplies
basic interface elements such as buttons and text fields.
\textbf{04 Media} supplies images, logos, illustrations, and icons with
stable identifiers. Designers verify that the required screens and
navigation references resolve, that listed assets are available, and
that the brief can be satisfied by multiple visual solutions.
Reference layouts and layout-specific decorative elements are withheld.
Figure~\ref{fig:sup_foodgo_starter} shows what agents receive for Foodgo.
Complete page captures appear in Sec.~\ref{sec:sup_foodgo_starter}.

\paragraph{Example: Foodgo.}
The product-detail example in Fig.~\ref{fig:main-bench} illustrates how
shared product content, imagery, and selection options admit different
designs. One output presents the image separately and arranges options
in compact grids. Another combines the image and description in a
hero card and uses horizontal option rows. The brief fixes the
content and required interactions while leaving the hierarchy,
component treatment, and arrangement open.

\subsection{Agent Execution Configuration}
\label{sec:sup_agent_config}

\noindent\textbf{Shared task environment.}
All eleven models are evaluated using the same Claude Code
agent harness and execution configuration. For each
task--model pair, we prepare an independent, writable
copy of the corresponding Figma Starter File to prevent
interference across runs. The file contains four reference
pages: \texttt{00 Start Here}, \texttt{01 Foundations},
\texttt{02 Components}, and \texttt{04 Media}, which
provide the task specification, visual guidelines,
reusable components, and design assets, respectively.
Agents construct their deliverables in a dedicated
\texttt{Workspace} page while preserving the reference
materials. The task specification fixes the required
screens, functional relationships, content, and data,
but leaves layout, visual hierarchy, component
composition, and visual presentation to the agent.

\noindent\textbf{Agent instructions.}
All models receive a shared instruction template
consisting of system-level instructions appended to
the Claude Code harness and a user task prompt.
Run-specific file URLs, file keys, and account
identifiers are substituted without changing the
underlying task protocol. Product-specific requirements
are obtained directly from the assigned Figma file.
The instruction templates are provided below, with
run-specific identifiers anonymized and page names
standardized.

\begin{promptbox}{Prompt 1: System-level Instructions}
You are completing an initial Figma product design
benchmark.

Assigned file: [FIGMA_URL] (fileKey [FILE_KEY]).
Use only this assigned file for the task. The user
has provided a dedicated writable benchmark copy.

Call whoami before writing and confirm the assigned
account is [FIGMA_ACCOUNT]. Use the configured account
and follow the live tool schemas; pass only arguments
those schemas accept.

Read 00 Start Here for the product, exact screens,
states, data, copy and prototype requirements.
If an app name in the external query disagrees with
this brief, use the product specified in 00 Start Here.

Read 01 Foundations for the design rules, including
the required fonts and colors. Reuse the components
in 02 Components. Read and reuse the assets in
04 Media. Preserve all reference pages and their
existing nodes.

Create the complete design inside Workspace in the
assigned file. If this page is missing, create it.
New frames, reusable components and prototype
connections are allowed in this workspace, including
when it starts empty.

For this benchmark, starter instructions about
keeping the source read-only or delivering in
another new file apply to preserving the reference
pages, not to the workspace. The user's delivery
location is Workspace in this assigned copy.

Keep existing unrelated content intact. Do not
replace the reference pages, create a different
Figma file or switch accounts or teams to complete
the task.

Use whoami -> get_metadata and read-only use_figma
to inspect the reference pages -> use_figma to build
the workspace -> get_screenshot to verify the designs.
Use the supplied Figma tools directly, without
delegating the design to other agents.

Follow the visual requirements in the file. Fonts,
gradients, shadows and blurs are allowed when
supported by the Figma API and appropriate to those
requirements; do not substitute an unrelated
universal font or style restriction.

Use valid Figma Plugin API calls and load the
required fonts before editing text. Build in
manageable sections and check tool results before
continuing. Add the prototype interactions required
by the brief after the relevant screens exist.

Inspect screenshots of all required screens.
You may take multiple screenshots, fix visible
problems and check again. A successful screenshot
alone is not task completion: finish the required
screens, states and prototype flows before the
final answer.

This is an unattended session. Resolve routine
implementation choices using the brief and continue
without waiting for a human answer. If permissions
or quota prevent progress, report the actual
limitation and the unfinished work.

Finish with the assigned file URL, the Workspace
page ID, the screen IDs, the prototype work completed
and any remaining limitations. Do not claim
completion for work you did not finish.
\end{promptbox}

\noindent\textit{User task prompt template.}
The user prompt directs the agent to the assigned
Figma file and its task resources. The same English
template is used across models, with the file URL
replaced for each run.

\begin{promptbox}{Prompt 2: User Task Prompt}
Please complete the full design in the following
Figma file:

Figma file: [FIGMA_URL]

First, read the design requirements in
"00 Start Here", refer to the visual guidelines
in "01 Foundations", reuse the components in
"02 Components", and use the assets in
"04 Media". Deliver the complete design in
"Workspace".
\end{promptbox}

\noindent\textbf{Tool access and native editing.}
The Claude Code harness exposes four Figma MCP tools:
\texttt{whoami}, \texttt{get\_metadata},
\texttt{get\_screenshot}, and \texttt{use\_figma}.
These support account verification, structural
inspection, visual feedback, and native design
manipulation, respectively. In particular,
\texttt{use\_figma} executes JavaScript within the
Figma Plugin API environment, enabling direct
creation and modification of frames, text,
components, and prototype connections. Agents
are restricted to the assigned Figma file and
the designated tool interface. Auxiliary agent-side
tools, external code execution, and HTML/CSS-to-Figma
conversion pipelines are excluded.

\noindent\textbf{Execution and artifact collection.}
All models follow the same execution controls,
including a 5,400-second time limit per run.
An execution terminates when the agent submits
its final response or when the execution
controller reaches a stopping condition.
During generation, agents may inspect intermediate
screenshots and iteratively refine their designs.
After execution, the evaluation platform independently
inspects the assigned Workspace, identifies the
generated frames, and checks their coverage against
the required screen inventory. Rendered screenshots
and Figma layer summaries are then collected for
subsequent evaluation. Before scoring, rule-based checks
verify the required screen count and inspect rendered
screen images for blank content. Only artifacts that
pass both checks proceed to scoring.

The platform records execution status, elapsed
time, tool activity, and artifact identifiers.
Interrupted attempts and subsequent retries are
tracked separately, with evaluation results linked
to the corresponding artifact version. Execution
completion and required-screen coverage are
distinguished from design quality, which is
assessed independently through the rubric-based
and human-preference evaluation protocols.


\section{Evaluation Protocols}
\label{sec:sup_evaluation}

\subsection{Complete Rubric Definitions}
\label{sec:sup_rubrics}

Table~\ref{tab:evaluation_rubrics} summarizes the criterion counts and
maximum scores by dimension.

We evaluate each generated Figma design using version 4 of a fixed rubric with 22 scored criteria. The criteria cover two perspectives: usability and ease of use (A, B, G; 27 points) and aesthetics (D, E, F; 26 points).
The six dimensions, using the terminology of the main paper, are user experience (A1--A4, 8 points), information organization (B1--B5, 15 points), deliverable quality (G1--G2, 4 points), layout quality (D1--D4, 12 points), consistency (E1--E2, 4 points), and visual presentation (F1--F5, 10 points).
B1--B5 and D1--D4 use integer scores from 0 to 3. The other thirteen criteria use integer scores from 0 to 2. The item-specific anchors below govern scoring.

\paragraph{Pre-scoring checks.}
Scoring begins only after the generated artifact passes the rule-based
screen-count and blank-screen checks described in
Sec.~\ref{sec:sup_agent_config}. Artifacts that fail either check are
not scored; they must pass both checks after correction or regeneration
before entering evaluation. For an eligible artifact, the total score
$S$ is the sum of the 22 criterion scores, with a maximum of 53.
A percentage score is reported as $100S/53$.

\paragraph{Evidence and scoring procedure.}
The judge reviews all supplied screenshots in sequence before scoring, using the task brief for context and for identifying required core screens. Each item receives a score and a brief rationale identifying the screen, region, observed evidence, and its effect on the user or design. The original Chinese prompt asks for 30--120 Chinese characters per item. Scoring follows impact, and a defect is counted under multiple items only when it has independently observable effects. Applicable aspects are assessed within each item. For example, a workflow without a step indicator can receive full credit on A2 when that indicator is unnecessary. Dynamic behavior that cannot be established from screenshots is marked unverified in the rationale. When A3 is judged without realistic content, its conclusion is limited to the supplied sample. The output contains an integer for every item. G1 and G2 use the Figma layer summary, with a score of 1 for each affected item and an explicit missing-evidence note if the summary is unavailable or inadequate.

\begingroup
\newcommand{\rubricdimension}[2]{%
  \par\addvspace{\medskipamount}%
  \noindent\colorbox{cvprblue!36}{%
    \parbox{\dimexpr\linewidth-2\fboxsep\relax}{%
      \raggedright\color{cvprblue!52!black}\rmfamily\textbf{#1} (#2)}}%
  \par\nobreak
}
\newenvironment{rubriccard}[1]{%
  \par\addvspace{\medskipamount}%
  \noindent\colorbox{cvprblue!18}{%
    \parbox{\dimexpr\linewidth-2\fboxsep\relax}{\raggedright\color{cvprblue!42!black}\bfseries #1}}%
  \par\nobreak\vskip 1pt\nobreak
  \leftskip=\fboxsep\relax
  \rightskip=\fboxsep\relax
  \noindent\ignorespaces
}{%
  \par\nobreak\vskip 1pt\nobreak
  {\color{cvprblue!32}\hrule}%
}
\newcommand{\rubricscore}[1]{%
  \par\noindent\hangindent=1.3em\relax
  \makebox[1.3em][l]{\color{cvprblue!52!black}\rmfamily\bfseries #1}%
}

\subsubsection*{Usability and Ease of Use}

\rubricdimension{User Experience}{A1--A4, 8 points}
\begin{rubriccard}{A1. Immediate understanding of the screen (0--2)}
Assess whether the title, principal content, and key labels convey the screen's purpose, primary subject, and current task. B3 covers logical correctness and A3 covers prominence.
\rubricscore{0} A key screen is unclear, incorrect, or contradictory on one of these aspects and likely to mislead.
\rubricscore{1} Some cues are incomplete, but the user can infer meaning from other information on the screen.
\rubricscore{2} The purpose, subject, and current task are directly apparent on each key screen.
\end{rubriccard}

\begin{rubriccard}{A2. Orientation and next steps (0--2)}
Assess location, workflow step, selection, system status, and the expected next step or exit where relevant. A3 covers findability and A4 covers control recognition.
\rubricscore{0} Essential state is absent or incorrect, or a necessary next step or exit is unclear, causing misjudgment or interrupting the workflow.
\rubricscore{1} Some cues are unclear, but context supports the main workflow.
\rubricscore{2} Location and state are clear, and the main workflow's next steps and exits are apparent.
\end{rubriccard}

\begin{rubriccard}{A3. Findability of core content and actions (0--2)}
Assess whether users can locate core content and primary actions at the observed information density, with minimal interference from secondary material. B2 covers ordering and B5 covers relative emphasis within an action set.
\rubricscore{0} Core content or a primary action is absent, hidden, or substantially obscured, impairing the main task.
\rubricscore{1} Local distractions or weak emphasis require brief searching, with the main task intact.
\rubricscore{2} Core content and primary actions are readily found on key screens, with little interference from secondary material.
\end{rubriccard}

\begin{rubriccard}{A4. Recognizability and meaning of controls (0--2)}
Assess whether controls look actionable and whether icons, labels, and field instructions convey the intended action or input. A3 covers findability, B4 covers icon semantics in detail, and D1 and D2 cover layout. Hit-target size falls outside this item.
\rubricscore{0} Key controls are unrecognizable or misleading, or key input requirements are unclear enough to cause errors.
\rubricscore{1} A few controls or instructions require contextual inference, with core actions intact.
\rubricscore{2} Primary controls are recognizable, their icons and labels communicate their purpose, and key input requirements are clear.
\end{rubriccard}

\rubricdimension{Information Organization}{B1--B5, 15 points}
\begin{rubriccard}{B1. Information grouping (0--3)}
Assess whether grouping by task, entity, or category reflects meaningful relationships and uses suitable granularity. Alignment, spacing, and whitespace are layout criteria.
\rubricscore{0} Core content is so mixed, misgrouped, or fragmented that its membership or relationships cannot be determined reliably.
\rubricscore{1} Key groups or multiple groups require repeated searching and reconstruction.
\rubricscore{2} A few groups mix or split related content, or use uneven granularity, but overall relationships remain clear.
\rubricscore{3} Groups are coherent, related information stays together, and granularity suits the content.
\end{rubriccard}

\begin{rubriccard}{B2. Information sequencing (0--3)}
Assess whether ordering supports reading, decision-making, and workflow progression, with key information presented when needed. Visual prominence, alignment, and module spacing are scored separately.
\rubricscore{0} Key premises, conclusions, or steps are ordered so poorly that users cannot understand or advance the main task.
\rubricscore{1} Key information arrives conspicuously late, or disconnected content requires frequent skipping and backtracking.
\rubricscore{2} Some ordering warrants brief backtracking, with the overall sequence still easy to follow.
\rubricscore{3} Ordering supports reading and decisions, key content appears first where appropriate, and related information follows naturally.
\end{rubriccard}

\begin{rubriccard}{B3. Logical consistency of information and workflow (0--3)}
Assess correctness and consistency of information, data, states, available actions, and domain relationships within and across related screens. Limit judgments to facts that can be established from the brief or design.
\rubricscore{0} A core contradiction prevents a reliable judgment or leads to a wrong conclusion or action.
\rubricscore{1} A key stage or multiple parts are inconsistent enough that users must reconcile contradictions themselves.
\rubricscore{2} Minor inconsistencies leave the main information and domain relationships understandable.
\rubricscore{3} Information, states, actions, and relationships are coherent, with no identifiable contradictions or implausible semantics.
\end{rubriccard}

\begin{rubriccard}{B4. Icon legibility and semantics (0--3)}
Assess whether icons are visually distinguishable and quickly communicate their function or status in context. Clickability, stylistic consistency, and drawing quality are separate concerns.
\rubricscore{0} A key icon misrepresents its meaning and is likely to prompt an incorrect interpretation or action.
\rubricscore{1} One key icon or several icons remain ambiguous even in context.
\rubricscore{2} A few icons are unintuitive but understandable with nearby labels or context.
\rubricscore{3} Icons are legible and communicate the intended function or status quickly and accurately in context.
\end{rubriccard}

\begin{rubriccard}{B5. Visual priority of actions (0--3)}
Assess relative emphasis of primary, secondary, and destructive actions within the same action set, given task priority and risk. Page-wide findability, state-colour semantics, and safeguards such as confirmation dialogs fall outside this item.
\rubricscore{0} The hierarchy presents a destructive or incorrect action as the apparent default, making accidental selection highly likely.
\rubricscore{1} The primary action is too weak, actions compete for attention, or destructive actions are hard to distinguish.
\rubricscore{2} Local emphasis is uneven, but primary and destructive actions remain identifiable.
\rubricscore{3} The hierarchy clearly guides the intended action, and destructive actions are recognizable without inviting accidental selection.
\end{rubriccard}

\rubricdimension{Deliverable Quality}{G1--G2, 4 points}
\begin{rubriccard}{G1. Naming of key Figma layers (0--2)}
Assess meaningful, accurate, and consistent naming of pages, frames, components, and key modules. Individual decorative layers and ordinary vectors need no custom names.
\rubricscore{0} Most pages use default or meaningless names for key structures, preventing navigation by name.
\rubricscore{1} Most key structures can be found, with occasional default, ambiguous, or inconsistent names.
\rubricscore{2} Major pages, frames, components, and modules are clearly and consistently named and easy to locate.
\end{rubriccard}

\begin{rubriccard}{G2. Figma layer hierarchy (0--2)}
Assess whether parent--child relationships reflect functional regions and actual containment.
\rubricscore{0} Hierarchy is ineffective, frames are empty, or many elements sit directly under the page or root frame.
\rubricscore{1} Major functional modules are grouped, but their contents are largely flat.
\rubricscore{2} Key layers belong to the correct functional regions, and modules have meaningful internal nesting.
\end{rubriccard}

\subsubsection*{Aesthetics}

\rubricdimension{Layout Quality}{D1--D4, 12 points}
\begin{rubriccard}{D1. Layout within a single control (0--3)}
Assess proportions, alignment, spacing, and padding among the text, icon, and container of an individual control. D2 covers arrangement across elements in a composite block. Semantics and hit-target size fall outside D1.
\rubricscore{0} Severe overlap, clipping, or misalignment within a key control obscures content or prevents use.
\rubricscore{1} Key controls or many controls show conspicuous imbalance, displacement, or crowding.
\rubricscore{2} Local proportion, alignment, or spacing issues leave content clear and operation largely unaffected.
\rubricscore{3} Internal proportions, alignment, spacing, and padding are balanced and keep content clear.
\end{rubriccard}

\begin{rubriccard}{D2. Layout within a composite block (0--3)}
Assess how elements are arranged, aligned, and spaced inside a functional or content block, such as a card, dialog, or complex form row. Consider padding and how spacing communicates grouping and priority.
\rubricscore{0} Severe overlap, occlusion, or displacement makes a key block's elements and relationships unusable or unintelligible.
\rubricscore{1} Disorganized, crowded, or overly dispersed elements obscure grouping or priority in a key block or multiple groups.
\rubricscore{2} Local alignment, spacing, or hierarchy issues leave grouping and relationships recognizable.
\rubricscore{3} Elements align and read in order, spacing distinguishes groups, and grouping and priority are clear.
\end{rubriccard}

\begin{rubriccard}{D3. Layout between major screen modules (0--3)}
Assess proportions, placement, alignment, and spacing of independent components, cards, or modules across the screen.
\rubricscore{0} Severe overlap, occlusion, or displacement breaks the page structure and prevents browsing core content.
\rubricscore{1} Major regions or the overall page are visibly imbalanced, with crowding or separation that obscures sections and the reading path.
\rubricscore{2} Local module layout issues leave sections and the reading path clear.
\rubricscore{3} Modules have suitable placement and proportions, coordinated alignment and spacing, and a clear reading path.
\end{rubriccard}

\begin{rubriccard}{D4. Excess unused space (0--3)}
Assess large purposeless gaps that interrupt structure or increase browsing effort. Allow intentional whitespace, natural unused space on short screens, and reserved areas with a clear purpose. D3 covers ordinary spacing between modules.
\rubricscore{0} Unused space dominates the screen and separates or displaces core content enough to prevent effective browsing.
\rubricscore{1} Large gaps in key or multiple areas cause clear discontinuities or substantial extra scrolling.
\rubricscore{2} A localized gap makes the screen feel sparse or adds some scrolling, but the main content remains continuous.
\rubricscore{3} Whitespace suits the amount of content and the screen structure, with no large purposeless gaps.
\end{rubriccard}

\rubricdimension{Consistency}{E1--E2, 4 points}
\begin{rubriccard}{E1. Consistency of equivalent UI elements (0--2)}
Assess whether equivalent controls, content units, and states share naming, structure, styling, and behavior where observable. Distinct functions may use distinct conventions.
\rubricscore{0} Key conventions conflict or equivalent elements vary widely, causing misrecognition or relearning.
\rubricscore{1} A few differences leave recognition and expectations intact.
\rubricscore{2} Equivalent elements follow consistent conventions, and equivalent actions and states are predictable.
\end{rubriccard}

\begin{rubriccard}{E2. Consistency across screens (0--2)}
Assess stability of navigation, layout framework, text hierarchy, colour roles, and visual language across screens. Allow differences driven by the workflow. E1 covers local elements and F5 covers overall polish.
\rubricscore{0} Conflicting conventions across key or multiple screens require users to relearn navigation, structure, or visual meaning.
\rubricscore{1} A few screens deviate locally while the overall framework remains consistent.
\rubricscore{2} Screens share overall conventions, so expectations transfer across screens.
\end{rubriccard}

\rubricdimension{Visual Presentation}{F1--F5, 10 points}
\begin{rubriccard}{F1. Typography and reading comfort (0--2)}
Assess font size and weight, line spacing, line length, wrapping, and their combined effect on hierarchy and readability. Content correctness and cross-screen consistency are separate concerns.
\rubricscore{0} Confused hierarchy, pervasive crowding, unsuitable line lengths, or broken wrapping materially impedes reading.
\rubricscore{1} Local hierarchy, density, or wrapping issues leave overall reading largely unaffected.
\rubricscore{2} Hierarchy is clear, density and line lengths are suitable, and text wraps naturally.
\end{rubriccard}

\begin{rubriccard}{F2. Colour roles and emphasis (0--2)}
Assess the roles of primary, secondary, and neutral colours and how emphasis and background choices guide attention. E2 covers cross-screen consistency, F3 state semantics, and F4 hierarchy among neutrals.
\rubricscore{0} Confused colour roles or excessive emphasis obscure key content or persistently draw attention to secondary material.
\rubricscore{1} Local overemphasis, weak emphasis, or distracting backgrounds leave the main focal points discernible.
\rubricscore{2} Colour roles are clear, emphasis is restrained, and backgrounds support focal content.
\end{rubriccard}

\begin{rubriccard}{F3. Semantic state colours (0--2)}
Assess colours for success, warning, error, danger, and disabled states against common semantics or explicit product conventions. Redundant non-colour cues and aesthetic appeal fall outside this item.
\rubricscore{0} Key colours conflict with their meanings, or multiple misleading choices are likely to cause misjudgment.
\rubricscore{1} Some colours are imperfect, but text or icons allow correct interpretation.
\rubricscore{2} State colours communicate their intended meanings clearly.
\end{rubriccard}

\begin{rubriccard}{F4. Tonal hierarchy of neutral colours (0--2)}
Assess how neutrals distinguish primary text, secondary text, supporting information, and separators while preserving readability. Brand and state colours and the size of the palette fall outside this item.
\rubricscore{0} Neutral tones obscure the hierarchy or make needed supporting content difficult to read.
\rubricscore{1} Some levels are too similar or too faint, but the hierarchy and content remain readable.
\rubricscore{2} Neutral tones distinguish levels clearly, and supporting information remains readable with suitable emphasis.
\end{rubriccard}

\begin{rubriccard}{F5. Overall visual polish (0--2)}
Assess coherence, refinement, and visual appeal across composition, colour, graphics, and details. Decorative quantity carries no credit by itself, and local defects should not be accumulated again here.
\rubricscore{0} Imbalanced composition, conflicting styles, or rough graphics and details prevent a coherent presentation.
\rubricscore{1} The overall design is coherent, but ordinary or underdeveloped composition, graphics, or details limit its refinement and appeal.
\rubricscore{2} Composition is balanced, colours and graphics work together, and the finished details contribute to a visually appealing whole.
\end{rubriccard}

\endgroup

\subsection{VLM-Based Evaluation Protocol}
\label{sec:sup_vlm}

\paragraph{Judges and evaluation evidence.}
We employ three VLM judges: Claude Opus 5.5, GPT-6.1 Sol,
and DeepSeek V4.1 Flash. Each judge independently evaluates
each submitted design artifact in three scoring passes,
yielding nine judgments per artifact. All passes use the
same frozen artifact version and evidence package, comprising
the task brief, an ordered inventory of screen images, the
corresponding screenshots, a Figma layer summary, and the
complete 22-criterion rubric. Screenshots provide visual
evidence, while the layer summary supplies structural
information, including layer naming, hierarchy, and node
statistics. Each judgment evaluates the complete multi-screen
deliverable rather than individual screens. The original
reference design is not used as a scoring target.

\paragraph{Prompt and structured output.}
The judge is instructed to inspect all supplied screens
and score all 22 criteria using their respective scoring
anchors. Each criterion requires an integer score and an
evidence-based rationale identifying the relevant screen
or region. The complete
criterion definitions and scoring anchors are provided
in Sec.~\ref{sec:sup_rubrics}.
\begin{tcolorbox}[
    enhanced,
    breakable,
    colback=black!2,
    colframe=black!35,
    colbacktitle=black!8,
    coltitle=black,
    fonttitle=\bfseries\small,
    title={VLM Judge Instructions},
    boxrule=0.5pt,
    arc=2pt,
    left=7pt,
    right=7pt,
    top=6pt,
    bottom=6pt
]
\small
\setlength{\parindent}{0pt}
\setlength{\parskip}{4pt}
\setlist[itemize]{
    leftmargin=*,
    itemsep=2pt,
    topsep=3pt,
    parsep=0pt
}
\setlist[enumerate]{
    leftmargin=*,
    itemsep=3pt,
    topsep=3pt,
    parsep=0pt
}

You are a senior product design reviewer. Evaluate the submitted
design using the fixed Version 4 evaluation rubric provided with
this request. The rubric contains 22 scored criteria and applies
uniformly to all submissions.

Your judgments must be grounded in the supplied evidence.
Apply the criteria consistently to support reproducible
evaluations across reviewers and evaluation sessions.
Do not modify the rubric or introduce additional criteria.

\par\medskip
\textbf{1. Review All Images Before Scoring}
\par\smallskip

This evaluation is conducted through a native multimodal API.
The screenshots listed in the screenshot manifest are embedded
in this request in the same numbered order.

Review the embedded images directly. No local filesystem,
Read tool, or Figma tool is available. Do not request or invoke
tools, and do not treat the absence of these tools as evidence
that a screenshot is missing. Any file paths in the manifest
serve only as identifiers.

Follow these requirements:
\begin{itemize}
    \item Inspect every screenshot individually, in the order
    specified in the manifest.

    \item Do not skip screenshots or infer the contents of
    later screenshots from earlier ones.

    \item Do not assign or output scores until you have
    reviewed all available screenshots.

    \item Treat a screenshot as unavailable only if it was
    not received or its content cannot be discerned.

    \item Report unavailable screenshots in
    \texttt{g0.reason} and identify the affected screens in
    the relevant criterion evidence, using the form
    \texttt{Screen N is unavailable}.
\end{itemize}

Base the evaluation only on the supplied screenshots,
design task, and layer structure summary. Use the design task
to establish requirements, but do not assume that a requested
feature is present unless the supplied evidence demonstrates it.

\par\medskip
\textbf{2. Apply G0 as a Prerequisite Gate}
\par\smallskip

G0 is a pass/fail gate, not a scored criterion.

Determine whether the generated design remains a usable basis
for further design work: its main content must be identifiable,
and it must not exhibit severe, widespread failures such as
overlapping content, misalignment, clipping, or missing core pages.

Fail G0 only when the design has fundamentally broken down.
Examples include:
\begin{itemize}
    \item The submission cannot be viewed, is almost entirely
    blank, or contains only isolated blocks of color.

    \item Severe, widespread overlap, misalignment, or clipping
    makes the main content unrecognizable or unusable.

    \item Core pages explicitly required by the task are
    extensively missing, so the main workflow cannot be assessed.

    \item The supplied layer evidence confirms that an entire
    page or a substantial functional region has been flattened
    into a raster image in the source design.
\end{itemize}

The use of screenshots as evaluation inputs does not establish
that the source design has been flattened. Do not infer this
condition from screenshots alone.

Minor defects, weak visual appeal, or the absence of a secondary
page do not, by themselves, justify failing G0. Assess such issues
under the relevant scored criteria.

If G0 fails, the system will set the aggregate score to 0.
You must still return all 22 criterion entries in the required
format. Do not calculate or output the aggregate score.

\par\medskip
\textbf{3. Apply the Fixed Rubric Consistently}
\par\smallskip

Apply the criterion definitions, scope boundaries, and scoring
anchors in the provided rubric.

The rubric covers two dimensions:
\begin{itemize}
    \item \textbf{Usability:}
    User Experience (A), Information Organization (B),
    and Design Deliverability (G).

    \item \textbf{Aesthetics:}
    Layout Quality (D), Consistency (E),
    and Visual Expression Quality (F).
\end{itemize}

Assign only integer scores within each criterion's permitted range:
\begin{itemize}
    \item B1--B5 and D1--D4: 0, 1, 2, or 3.
    \item All other scored criteria: 0, 1, or 2.
\end{itemize}

Apply the following evaluation rules:
\begin{enumerate}
    \item \textbf{Use the full scoring range when justified.}
    Do not default to intermediate scores. When the evidence
    is sufficient and no concrete issue can be identified
    within a criterion's scope, award the maximum score.
    Assign 0 when an observed failure meets that criterion's
    zero-score anchor.

    \item \textbf{Evaluate only applicable aspects of each criterion.}
    For example, assess workflow steps and selections under A2
    only when the scenario requires them. The absence of a step
    indicator is not inherently a defect. This rule does not
    permit omitting criterion entries from the output.

    \item \textbf{Score by impact, not by counting defects.}
    A single issue that prevents a critical judgment or task
    may justify 0. An inconsequential imperfection should not
    automatically reduce an otherwise justified score.

    \item \textbf{Avoid duplicate penalties.}
    Difficulty locating an action primarily belongs to A3;
    difficulty understanding its function after locating it
    primarily belongs to A4. Penalize both only when the
    evidence demonstrates two distinct effects. Follow the
    scope specified for each criterion when attributing issues.
    C1 and C2 are retired identifiers and must not be scored
    separately.

    \item \textbf{Distinguish insufficient evidence from failure.}
    Do not assign 0 solely because a property cannot be verified.
    When static screenshots cannot establish dynamic behavior,
    such as loading feedback or the result of a click, state
    \texttt{Not verified} in the relevant evidence and assess
    only what the available evidence supports.

    \item \textbf{Qualify conclusions based on limited content.}
    If realistic content is unavailable, explicitly state that
    the A3 judgment applies only to the supplied sample.

    \item \textbf{Attribute layout issues to the appropriate level.}
    Use D1 for issues within an individual control;
    D2 for issues within a functional block, such as a card,
    dialog, or form row;
    D3 for relationships between major page modules;
    and D4 for large areas of whitespace that lack a meaningful
    purpose.
\end{enumerate}

\par\medskip
\textbf{4. Assess G1 and G2 Using Layer Evidence}
\par\smallskip

Evaluate G1 and G2 from the supplied layer structure summary.
G1 concerns the naming of key structural elements;
G2 concerns parent--child relationships and functional containment.
These properties cannot be reliably determined from screenshots alone.

When interpreting the summary:
\begin{itemize}
    \item Evaluate names at the level of pages, frames,
    components, and key modules. Default names such as
    \texttt{Rectangle}, \texttt{Frame}, or \texttt{Group}
    are relevant when they prevent meaningful identification
    of these structures. Do not require every decorative or
    vector layer to have a descriptive name.

    \item Assess whether the hierarchy reflects the design's
    functional regions and containment relationships.
    A shallow hierarchy with many direct children may indicate
    insufficient grouping, but the score must follow the
    reported structure and the applicable scoring anchors.

    \item Do not introduce additional requirements, such as
    mandatory component-instance reuse, that are outside the
    defined scope of G1 and G2.
\end{itemize}

If the summary is missing or insufficient to assess either
criterion, assign that criterion a score of 1 and explicitly
state \texttt{Insufficient layer information} in its evidence.
This is the prescribed handling of missing evidence, not a
finding that the underlying structure meets the criterion.

\par\medskip
\textbf{5. Provide Specific Evidence and Return JSON Only}
\par\smallskip

For each criterion, provide an evidence string of
30--120 characters.

For screenshot-based judgments, identify:
\begin{itemize}
    \item The screen number.
    \item The specific region or element.
    \item The observable condition.
    \item Its consequence for understanding, task completion,
    or visual quality.
\end{itemize}

For G1 and G2, cite the relevant structural evidence and its
effect on design handoff. If the required layer information is
unavailable, use the missing-evidence statement specified above.

Describe concrete observations rather than generic assessments
such as \texttt{good}, \texttt{reasonable}, or \texttt{clear}
without supporting detail. Use the pattern
\textbf{observable evidence + consequence}.

Return one valid JSON object containing exactly two top-level
fields: \texttt{g0} and \texttt{scores}.

Output requirements:
\begin{itemize}
    \item \texttt{g0.pass} must be a Boolean.

    \item \texttt{g0.reason} must contain one sentence of
    no more than 40 characters explaining the gate decision.

    \item \texttt{scores} must contain exactly 22 entries,
    with each criterion appearing once.

    \item Each entry must contain exactly
    \texttt{key}, \texttt{score}, and \texttt{evidence}.

    \item Every score must be an integer within the
    criterion's permitted range.

    \item Use the following exact key order:

    \texttt{A1, A2, A3, A4, B1, B2, B3, B4, B5,
    G1, G2, D1, D2, D3, D4, E1, E2,
    F1, F2, F3, F4, F5}.

    \item Refer to screenshots as \texttt{Screen N}.
    Do not include file paths or filenames.

    \item Use standard JSON double quotes to delimit keys
    and string values. Do not include double-quotation
    characters within string values; use corner quotation
    marks (Unicode \texttt{U+300C} and \texttt{U+300D})
    when quoting interface text.

    \item Do not include Markdown, explanations outside
    the JSON object, additional fields, or an aggregate score.
\end{itemize}

The following example illustrates the structure only.
Replace the example values with evidence-based judgments
and include all 22 required entries in the actual response.

\begin{lstlisting}[
    basicstyle=\ttfamily\footnotesize,
    breaklines=true,
    breakatwhitespace=false,
    columns=fullflexible,
    keepspaces=true,
    showstringspaces=false,
    aboveskip=4pt,
    belowskip=4pt
]
{
  "g0": {
    "pass": true,
    "reason": "Main content is visible and usable."
  },
  "scores": [
    {
      "key": "A1",
      "score": 2,
      "evidence": "Screen 2, header: The title and active tab identify the page purpose and current task."
    }
  ]
}
\end{lstlisting}

\par\medskip
\textbf{6. Evaluation Inputs}
\par\smallskip

The following materials are provided as inputs for this evaluation.

\textbf{Evaluation Rubric}

\texttt{[EVALUATION\_RUBRIC]}

\textbf{Design Task}

\texttt{[QUERY]}

\textbf{Screenshot Manifest}

\texttt{[SCREENSHOT\_MANIFEST]}

\textbf{Layer Structure Summary}

\texttt{[LAYER\_STRUCTURE\_SUMMARY]}

\end{tcolorbox}

The response contains a \texttt{scores}
array containing \texttt{key}, \texttt{score}, and
\texttt{evidence} for each criterion. The parser verifies
complete and unique coverage of all
22 criteria, valid integer score ranges, and nonempty
evidence. Malformed or incomplete responses are retried
using the same evidence package. Responses that remain
invalid are recorded as technical failures rather than
quality judgments.

\paragraph{Artifact-level score aggregation.}
Let $x_{mtjrk}$ denote the score assigned to criterion $k$
for model $m$ on task $t$ by judge $j$ in pass $r$.
The score for each judgment and the ensemble score are

\begin{equation}
\begin{aligned}
S_{mtjr} &= \sum_{k=1}^{22}x_{mtjrk},\\
\bar{S}_{mt} &=
\frac{1}{9}\sum_{j=1}^{3}\sum_{r=1}^{3}S_{mtjr}.
\end{aligned}
\label{eq:sup_score}
\end{equation}

Each judgment has a maximum score of 53. The three judges
and three passes receive equal weight, and artifact-level
scores are normalized as $100\bar{S}_{mt}/53$.
No calibration against human preferences is applied.

For dimension $d$, let $\mathcal{K}_d$ denote its criteria
and $M_d$ its maximum score. The corresponding percentage is

\begin{equation}
Q_{mtd} =
\frac{100}{9M_d}
\sum_{j=1}^{3}\sum_{r=1}^{3}
\sum_{k\in\mathcal{K}_d}x_{mtjrk}.
\label{eq:sup_dimension}
\end{equation}

The six dimension maxima are 8, 15, 12, 4, 10, and 4,
following the order in Table~\ref{tab:evaluation_rubrics} of the main paper.
The overall percentage therefore represents a
maximum-score-weighted combination of dimension scores,
rather than their unweighted average. Model-level results
aggregate task-level scores, so tasks with more screens do
not receive greater weight solely due to screen count.

\paragraph{Validity and missing measurements.}
Only artifacts that pass the screen-count and blank-screen
checks are scored. Missing evaluation inputs, malformed
responses, and API failures are handled separately. Input completeness
and artifact-version consistency are checked before scoring.
Each judgment slot is retried when recoverable errors occur;
an official ensemble score is computed only when all nine
slots contain valid responses. For human--VLM comparisons,
both designs must have valid scores for the exact artifact
versions presented to human reviewers. Technical failures
and version mismatches are excluded from the corresponding
comparison denominator.

\subsection{Human Preference Evaluation Protocol}
\label{sec:sup_human_protocol}

\paragraph{Review unit and blinding.}
Professional designers compare two submissions for the same task. The interface provides the standardized task specification, aligned screen views, and access to the editable Figma artifacts, as described in Sec.~\ref{sec:design_evaluation}. Model identities and automated scores are concealed, and the left--right assignment is randomized. Reviewers assess the overall deliverable using requirement completeness, information organization, visual quality, and usability. An inability to judge is recorded separately rather than converted into a loss or a tie. Each overall observation is linked to its task, reviewer, review round, candidate identifiers, and artifact version.

\paragraph{Pair selection and within-task ordering.}
The Arena supports incremental pairwise sorting to avoid repeatedly
asking a reviewer to compare designs whose order is already known
within that review session. Direct comparisons establish partial
orders; subsequent comparisons merge ordered groups or insert a new
submission. Transitive relations can advance the sorting procedure,
but are stored separately from the submitted votes. Relations are
not transferred between reviewers, tasks, rounds, or artifact
versions. Cyclic or unresolved evidence requires review before a
complete task ordering is published. Corrected votes remain in the
history, while superseded observations are removed from the active
evidence. A completed ordering summarizes that review session; it
does not turn every inferred pair into a new independent human vote.

\paragraph{Win rates and preference ratings.}
For a set of eligible within-task observations, a model's win rate is
\begin{equation}
 \mathrm{WR}_m=\frac{W_m+\tfrac{1}{2}D_m}{W_m+L_m+D_m},
 \label{eq:sup_win_rate}
\end{equation}
where $W_m$, $L_m$, and $D_m$ denote wins, losses, and ties. Missing
or unjudgeable comparisons do not contribute to the denominator.
The Bradley--Terry model~\cite{bradley1952rank} expresses preference
between models $a$ and $b$ as
\begin{equation}
 P(a\succ b)=\frac{\exp(\theta_a)}
                  {\exp(\theta_a)+\exp(\theta_b)}.
 \label{eq:sup_bt}
\end{equation}
The strengths are estimated from within-task preference outcomes;
ties can contribute half an outcome to each direction. Elo-style
ratings are a display transformation,
$R_m=c+400\theta_m/\ln 10$, after fixing the additive offset of the
strength parameters. The constant $c$ selects a display baseline and
does not affect ranking or predicted preference probabilities.
Raw observed-vote win rates and win rates derived from completed
task orders have different sampling units and are not interchangeable.

\paragraph{Agreement on actual design comparisons.}
Let $\mathcal{V}$ contain eligible, still-effective overall human
votes for which both submitted artifact versions have valid rubric
scores. For a vote comparing $a$ and $b$ on task $t$, the machine
preference is the sign of
$\overline{S}_{at}-\overline{S}_{bt}$ from
Eq.~\ref{eq:sup_score}. We compare unrounded scores,
and equal scores produce a machine tie. Agreement is
\begin{equation}
 \mathrm{Agreement}=\frac{1}{|\mathcal{V}|}
  \sum_{v\in\mathcal{V}}
   \mathbf{1}[y_v^{\mathrm{human}}=y_v^{\mathrm{VLM}}].
 \label{eq:sup_pair_agreement}
\end{equation}
A tie agrees only with a tie. Each eligible overall vote is counted
once; inferred comparisons, screen-level choices, superseded votes,
and comparisons without a matching artifact version are excluded.
Separate reviewers or review rounds remain separate observations.
The denominator therefore reflects observed eligible judgments,
rather than all pairs that could theoretically be formed.

\paragraph{Agreement between model rankings.}
At the model level, we compare human-preference rankings with mean
rubric-score rankings on the aligned evaluation cohort. We use
Kendall's $\tau_b$ to account for ties:
\begin{equation}
 \tau_b=\frac{C-D}{\sqrt{(C+D+T_h)(C+D+T_q)}},
 \label{eq:sup_kendall}
\end{equation}
where $C$ and $D$ count concordant and discordant model pairs, and
$T_h$ and $T_q$ count pairs tied only in the human or rubric ranking,
respectively. Pairs tied in both rankings are omitted from these
counts. This statistic measures ordering across models, whereas
Eq.~\ref{eq:sup_pair_agreement} measures individual design comparisons;
the two metrics answer different questions.
\section{Fine-Tuning Details}
\label{sec:sup_ablation}
\subsection{Synthetic Defect Construction}
\label{sec:sup_defect_synthesis}

We construct 1{,}485 repair tasks from 846 source Figma screens
by injecting rubric-guided defects. Given an original screen, rule-based
miners identify eligible elements, and mechanism-specific mutators apply
executable edits to produce a corrupted screen. We define 20 defect
mechanisms spanning information organization, layout quality, consistency,
visual presentation, and deliverable quality. Each task contains multiple
defect instances, with weighted sampling prioritizing layout defects while
limiting the frequency of individual mechanisms. Compatibility checks
address node, subtree, and spatial conflicts, while geometric and layer-order
heuristics identify existing occlusion. Where applicable, perturbation
magnitudes are set to large or extra-large.

Table~\ref{tab:sup_defect_mechanisms} lists the 20 mechanisms and their
corresponding rubric criteria.
\begin{table}[!t]
  \centering
  \caption{\textbf{Synthetic defect mechanisms.} Criterion IDs correspond
  to the evaluation rubric.}
  \label{tab:sup_defect_mechanisms}
  \begingroup
  \small
  \setlength{\tabcolsep}{4pt}
  \renewcommand{\arraystretch}{1.08}
  \begin{tabularx}{\columnwidth}{@{}lX@{}}
    \toprule
    Criterion & Defect mechanism \\
    \midrule
    D3 & Misalignment \\
    E1 & Inconsistent peer styling \\
    D3 & Text occlusion \\
    D2 & Text overflow \\
    G1 & Generic layer names \\
    F4 & Low text contrast \\
    F1 & Undersized text \\
    D2 & Internal spacing drift \\
    F4 & Faint secondary text \\
    F1 & Flattened text hierarchy \\
    B5 & Weakened primary action \\
    D3 & Uneven spacing \\
    D2 & Inconsistent element sizes \\
    D4 & Empty region after removal \\
    B5 & Competing secondary action \\
    F1 & Reduced caption size \\
    D1 & Off-center control label \\
    B1 & Missing section title \\
    E1 & Missing component part \\
    B3 & Incorrect active tab \\
    \bottomrule
  \end{tabularx}
  \endgroup
\end{table}

\subsection{Repair-Based Fine-Tuning}
\label{sec:sup_repair_sft}

Given a corrupted Figma page and its editing scope, but no defect list,
Claude-Opus~5 acts as a teacher model and locates and repairs defects
in place through Figma MCP tool calls while preserving correct regions.
The resulting 1{,}485 multimodal trajectories include approximately
16{,}000 edit calls and 5{,}880 screenshot calls and record inspection,
editing, and visual verification. The training pages are disjoint from
the evaluation tasks. We split these trajectories into 1{,}470 training
samples and 15 validation samples and fine-tune all language-model
parameters of Qwen3.5-9B while freezing the vision encoder and multimodal
aligner, using the configuration in Table~\ref{tab:sup_sft_config}.

\begin{table}[!ht]
  \centering
  \caption{\textbf{SFT configuration for repair-based fine-tuning.}}
  \label{tab:sup_sft_config}
  \small
  \setlength{\tabcolsep}{4pt}
  \begin{tabularx}{\columnwidth}{@{}lX@{}}
    \toprule
    Parameter & Setting \\
    \midrule
    Base model & Qwen3.5-9B \\
    Training / validation samples & 1{,}470 / 15 \\
    Trainable parameters & All language-model parameters \\
    Frozen modules & Vision encoder and multimodal aligner \\
    Training epochs & 3 \\
    Training GPUs & 4 NVIDIA H800 \\
    Batch size per GPU & 1 \\
    Gradient accumulation steps & 1 \\
    Global batch size & 4 \\
    Maximum sequence length & 262{,}144 tokens \\
    Sequence packing & Disabled \\
    Optimizer & AdamW \\
    Learning rate & $1\times10^{-5}$ \\
    Learning-rate schedule & Cosine decay, 5\% warmup \\
    Adam $(\beta_1,\beta_2,\epsilon)$ & $(0.9,0.95,10^{-8})$ \\
    Weight decay & 0.1 \\
    Gradient clipping norm & 1.0 \\
    Precision & BF16 \\
    Distributed training & DeepSpeed ZeRO-2, optimizer CPU offload \\
    Random seed & 42 \\
    \bottomrule
  \end{tabularx}
\end{table}

\onecolumn
\section{Additional Qualitative Analysis}
\label{sec:sup_qualitative}

\subsection{Native Figma Structure Beyond Screenshots}
\label{sec:sup_native_structure}

\begin{figure}[!ht]
  \centering
  \includegraphics[width=\textwidth]{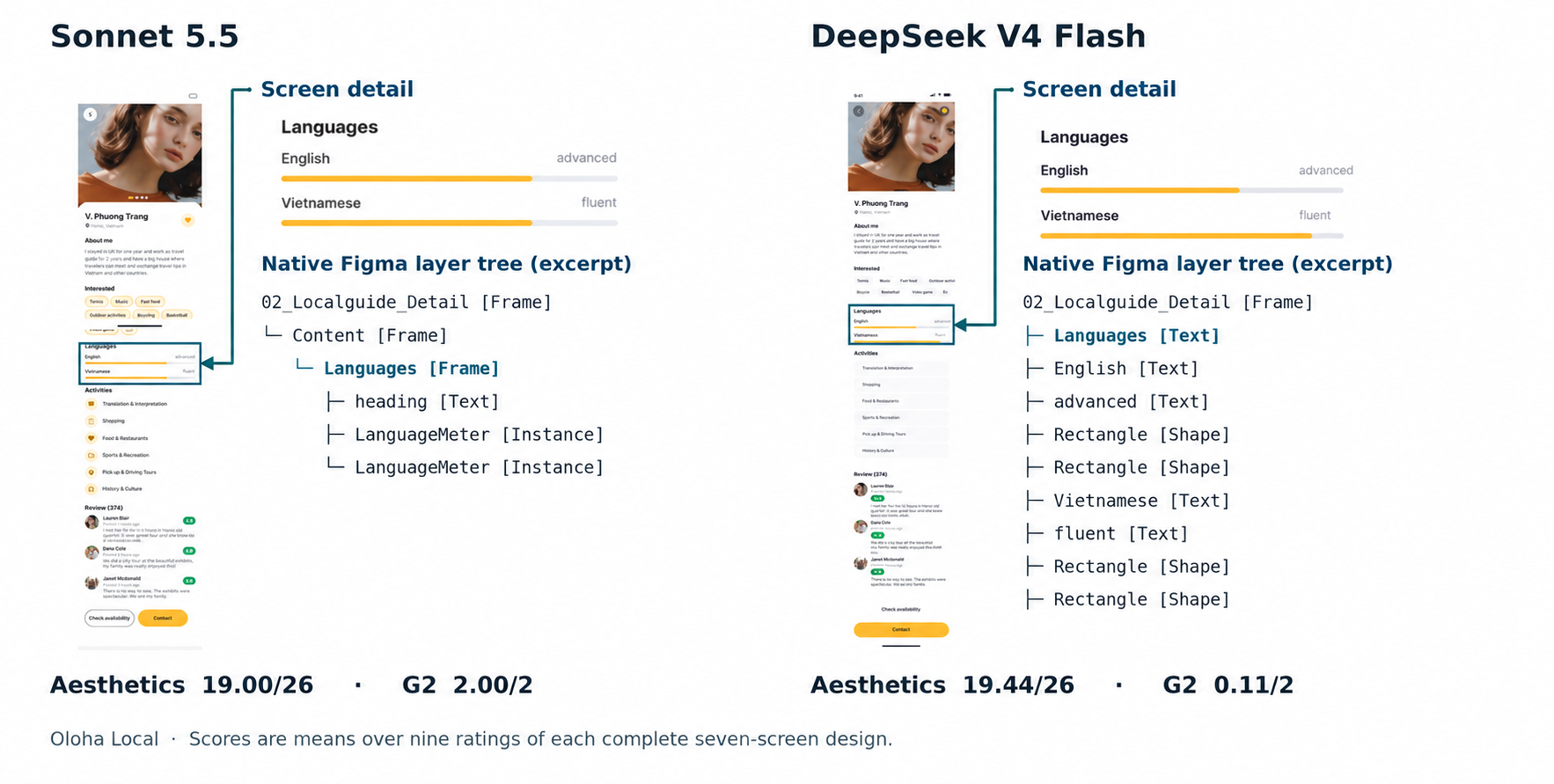}
  \caption{\textbf{Visual grouping and native layer hierarchy in Oloha Local.}
  The boxes locate the Languages section in each guide detail screenshot.
  The trees show parent--child relationships extracted from the original
  \texttt{.fig} files. Scores average nine ratings per complete
  seven-screen submission (three VLM judges, three runs each).}
  \label{fig:sup_oloha_native_structure}
\end{figure}

On the Oloha Local task, Claude Sonnet~5.5 and DeepSeek V4 Flash
receive similar aesthetics scores (19.00 and 19.44 out of 26, D+E+F).
Their layer hierarchy scores (G2) differ sharply, at 2.00 and 0.11
out of 2, respectively (Fig.~\ref{fig:sup_oloha_native_structure}).

Both guide detail screens visually group language names, proficiency
labels, and progress bars into one section. Sonnet uses a named
\texttt{Languages} frame with vertical Auto Layout to hold the heading
and two instances of the same \mbox{\texttt{LanguageMeter}} component.
DeepSeek places the heading, labels, and default-named bar rectangles
directly under the screen frame. An editor can locate Sonnet's section
by name and move it as a unit. Moving the same section in DeepSeek's
file requires selecting its constituent layers.

\subsection{Additional Failure Cases}

\label{sec:sup_failure_cases}

\paragraph{Failure Occurrence Statistics.}
To characterize recurring design failures, we analyze
the evaluation feedback obtained using our 22 rubrics
and group commonly observed issues into representative
failure categories. We then identify these failure
patterns from the rubric judges' textual feedback
using rule-based matching. For each evaluation run,
a failure category is counted once if the corresponding
issue is identified, regardless of how many times it
occurs. Multiple failure categories may co-occur
within the same run. The occurrence rate of each
category is calculated as the proportion of runs
exhibiting that failure. As shown in Fig.~\ref{fig:failure_modes},
excessive empty space (70\%), logic/data
inconsistencies (68\%), and misalignment or uneven
spacing (65\%) are among the most frequently
observed issues.

\paragraph{Representative Failure Examples.}
Figure~\ref{fig:failure_modes} illustrates empty space, data--visual
inconsistency, overlap, and overflow. For flat layers and layer naming,
see the native-file analysis in Sec.~\ref{sec:sup_native_structure}.
Section~\ref{sec:sup_failure_gallery} illustrates the remaining
categories in the order shown in the frequency chart, followed by a
separate example of a domain-specific visualization error.

\clearpage
\section{Complete Foodgo Starter File}
\label{sec:sup_foodgo_starter}

The following figures show all four pages of the Foodgo starter file.
The task specification spans two panels, read from left to right.

\begin{center}
  \captionsetup{hypcap=false}
  \begin{minipage}[t]{0.48\textwidth}
    \vspace{0pt}
    \includegraphics[width=\linewidth]{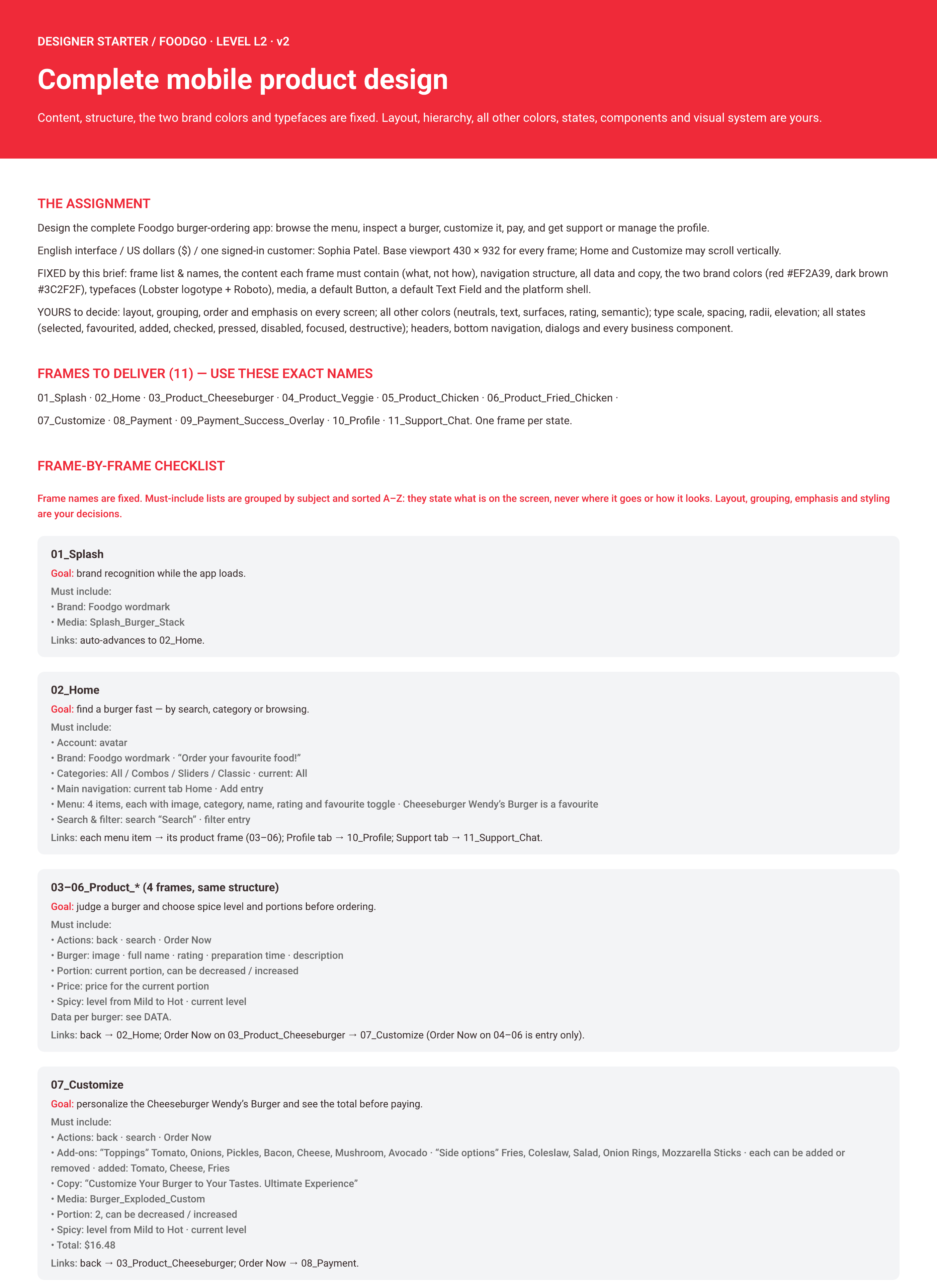}
  \end{minipage}\hfill
  \begin{minipage}[t]{0.48\textwidth}
    \vspace{0pt}
    \includegraphics[width=\linewidth]{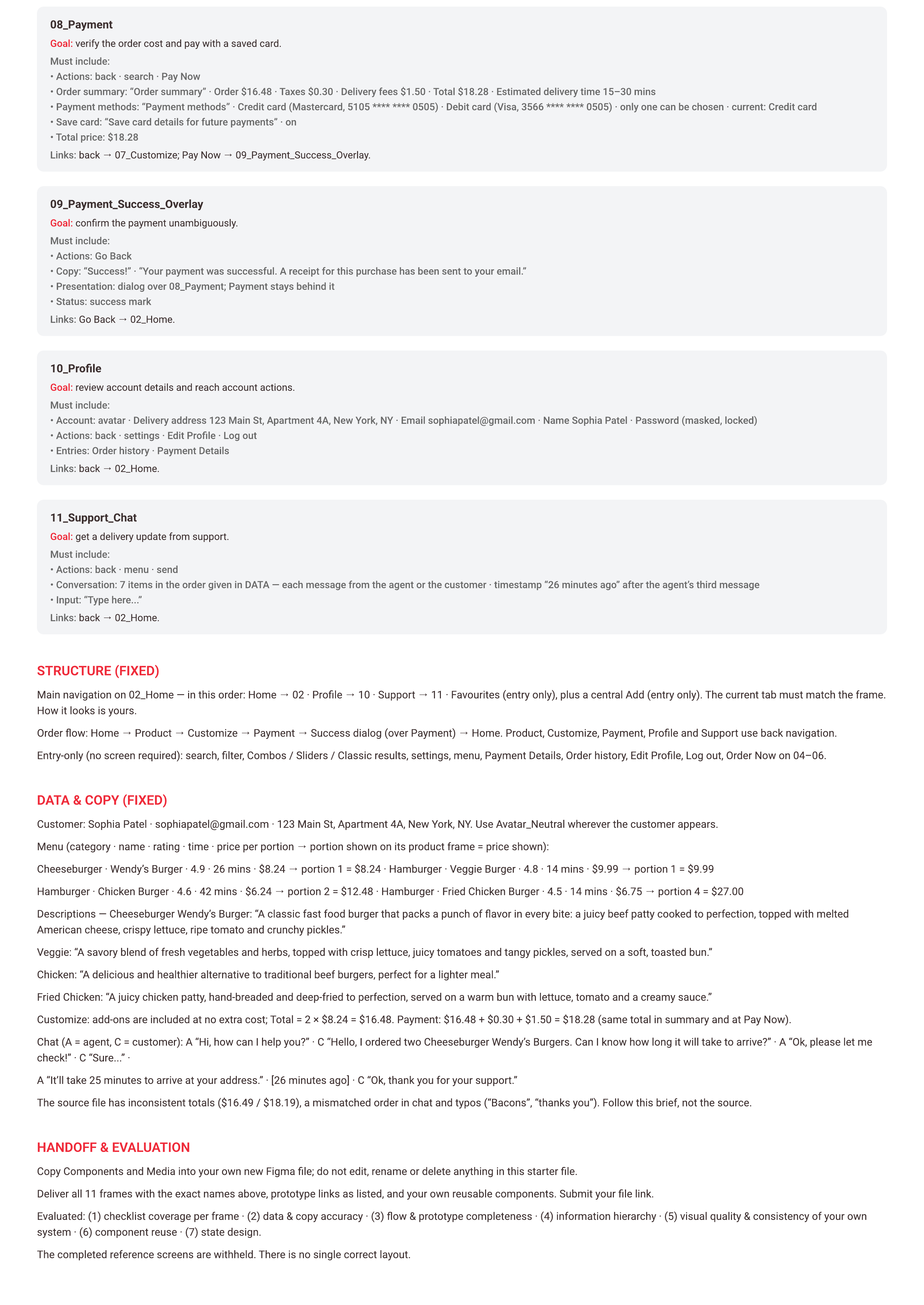}
  \end{minipage}
  \captionof{figure}{\textbf{00 Start Here.} The assignment and screen
  checklist begin on the left. The checklist continues on the right,
  followed by the workflow structure, fixed content, and handoff
  requirements.}
\end{center}
\clearpage

\begin{center}
  \captionsetup{hypcap=false}
  \includegraphics[width=0.88\textwidth]{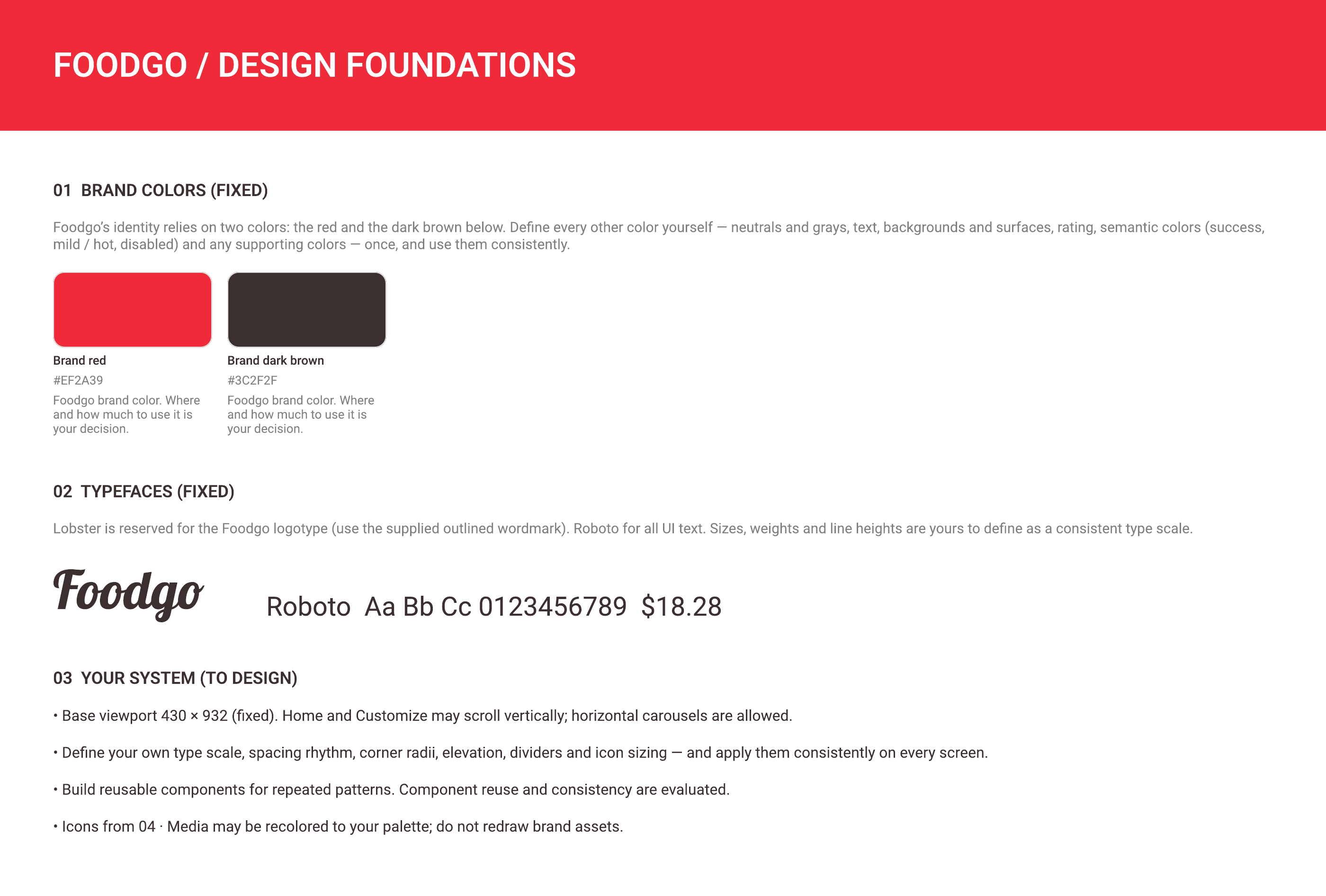}\par
  \medskip
  \includegraphics[width=0.88\textwidth]{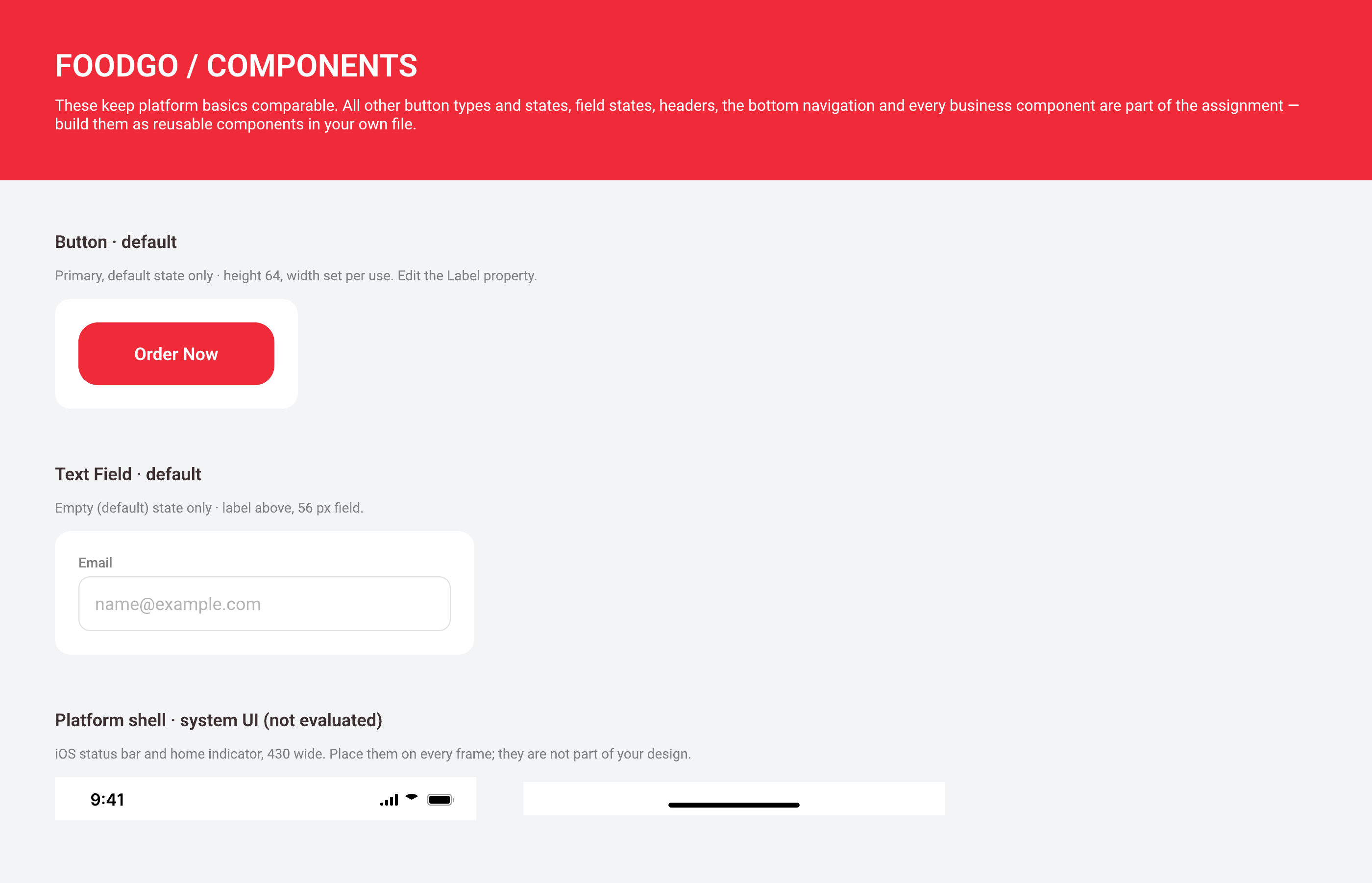}
  \captionof{figure}{\textbf{01 Foundations and 02 Components.} The upper page
  defines the brand colors and typefaces. The lower page supplies a
  default button, a default text field, and platform chrome.}
\end{center}
\clearpage

\begin{center}
  \captionsetup{hypcap=false}
  \includegraphics[width=\linewidth,height=0.85\textheight,keepaspectratio]{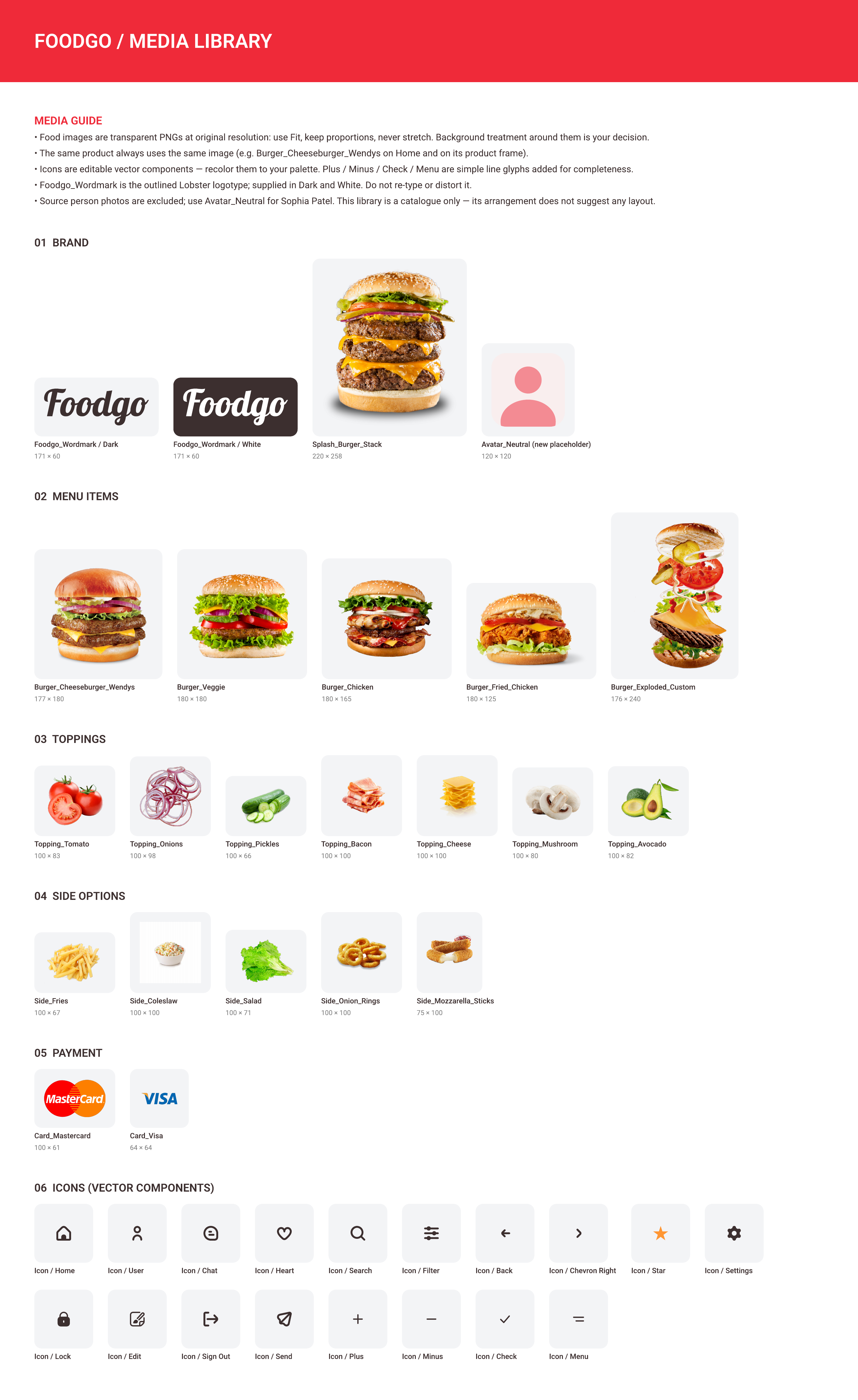}
  \captionof{figure}{\textbf{04 Media.} The library supplies brand, menu,
  topping, side, and payment assets, along with interface icons.}
\end{center}
\clearpage
\begingroup
\newcommand{\failurecrop}[6]{%
  \begin{minipage}[t]{#1\linewidth}
    \centering
    \vspace{0pt}%
    \includegraphics[width=\linewidth,height=#2,keepaspectratio,
      trim={#3},clip]{figures/failures/failure_examples/#4}\par
    \smallskip
    {\small #5\par\textbf{#6}\par}
  \end{minipage}%
}

\section{Representative Failure Examples}
\label{sec:sup_failure_gallery}

The headings follow the failure categories in Fig.~\ref{fig:failure_modes}.
Each excerpt retains the UI context needed to inspect the issue.
Each image credits the model that generated the design. Red outlines
are part of the original Figma exports.

\subsection{Misaligned / uneven spacing}
The hero image starts to the right of the article's text column.
In the dialog, the icon and confirmation button are centered while the
title is offset. The podcast list uses a common left edge, but its rows
end at different positions.

\begin{center}
  \failurecrop{0.39}{158bp}{0bp 88.40bp 0bp 0bp}
    {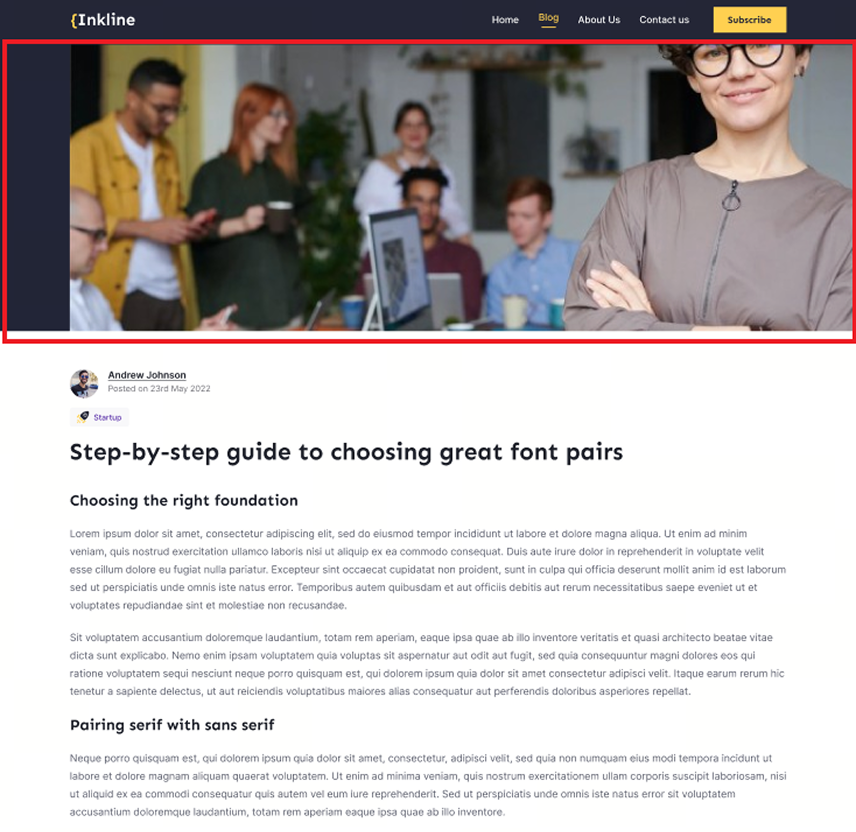}{Offset hero image}{Qwen 3.8 Max}\hfill
  \failurecrop{0.23}{158bp}{4.56bp 91.99bp 4.56bp 101.10bp}
    {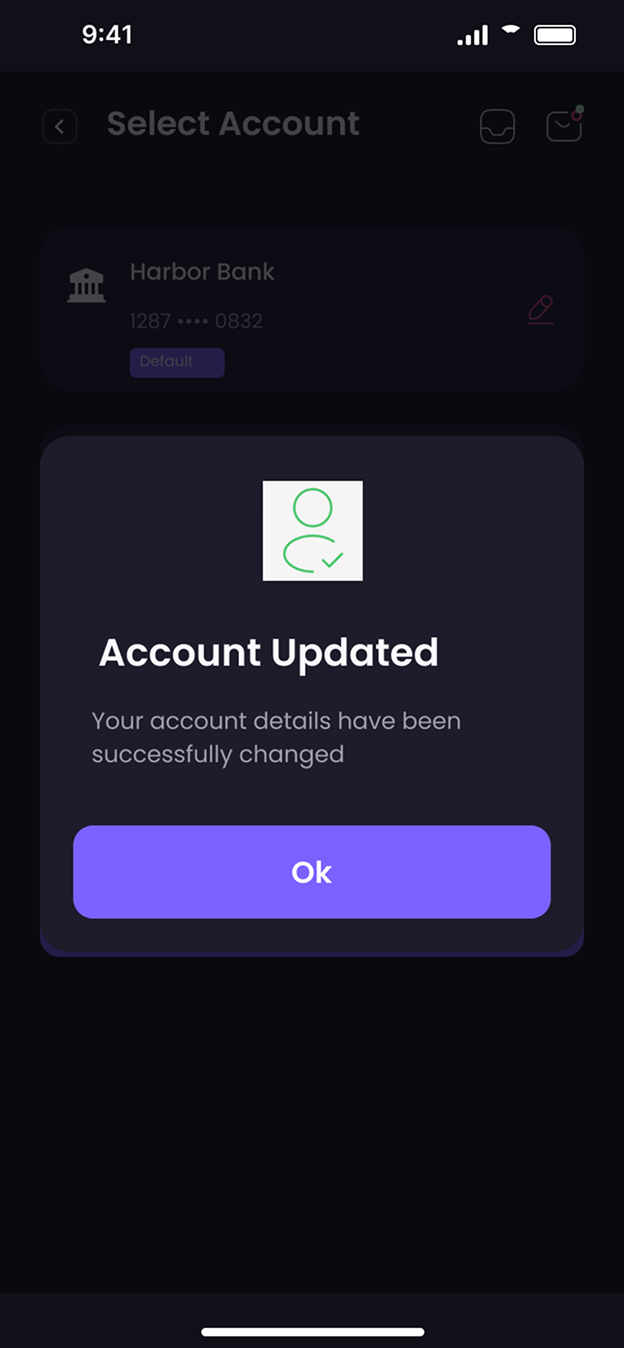}{Offset dialog title}{GPT-6.1 Sol}\hfill
  \failurecrop{0.34}{158bp}{20.82bp 141.12bp 124.90bp 70.56bp}
    {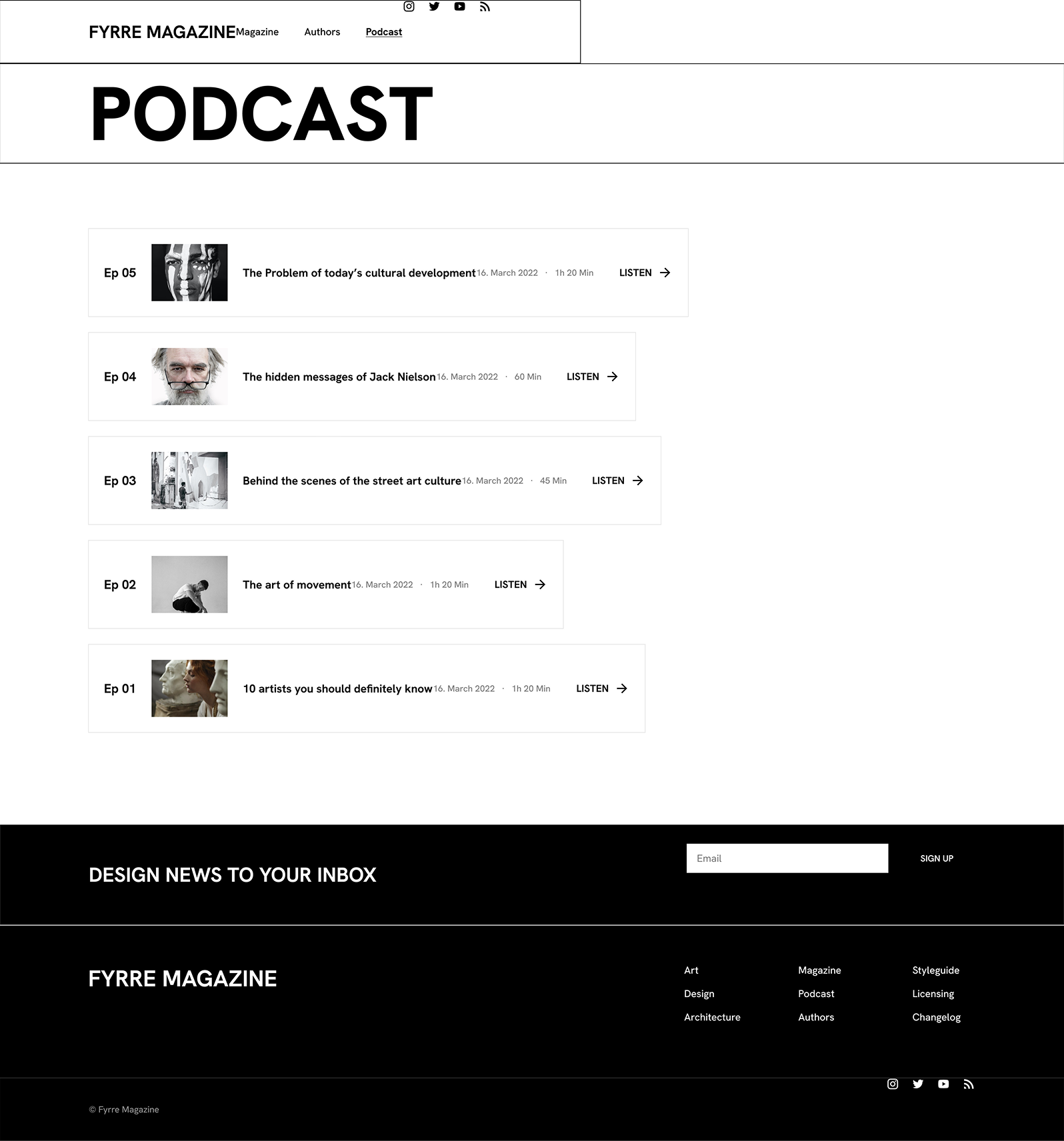}{Uneven podcast row widths}{Gemini 3.5 Flash}
\end{center}

\subsection{Inconsistent across screens}
The back button changes its circular fill and shadow treatment between
the Nordic Cottage and New Plan screens, although it serves the same
navigation role.

\begin{center}
  \failurecrop{0.80}{150bp}{0bp 247.91bp 0bp 0bp}
    {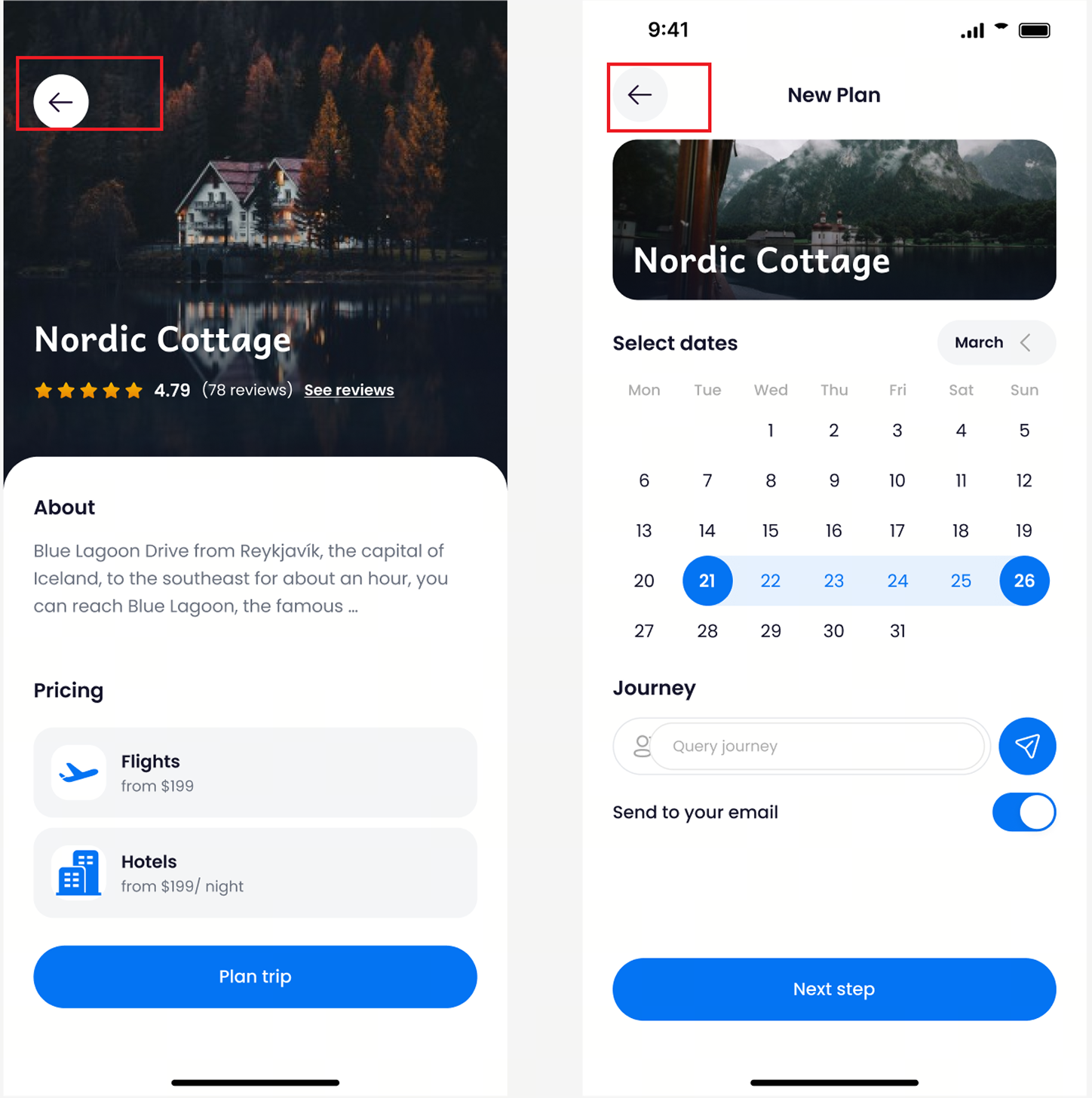}{Back buttons on two screens of the same workflow}{Claude Sonnet 5.5}
\end{center}

\clearpage

\subsection{Icon meaning unclear}
Both testimonial carousel controls use a right-pointing arrow, making
it difficult to distinguish the previous and next actions.

\begin{center}
  \failurecrop{0.64}{174bp}{258.81bp 143.64bp 24.44bp 208.28bp}
    {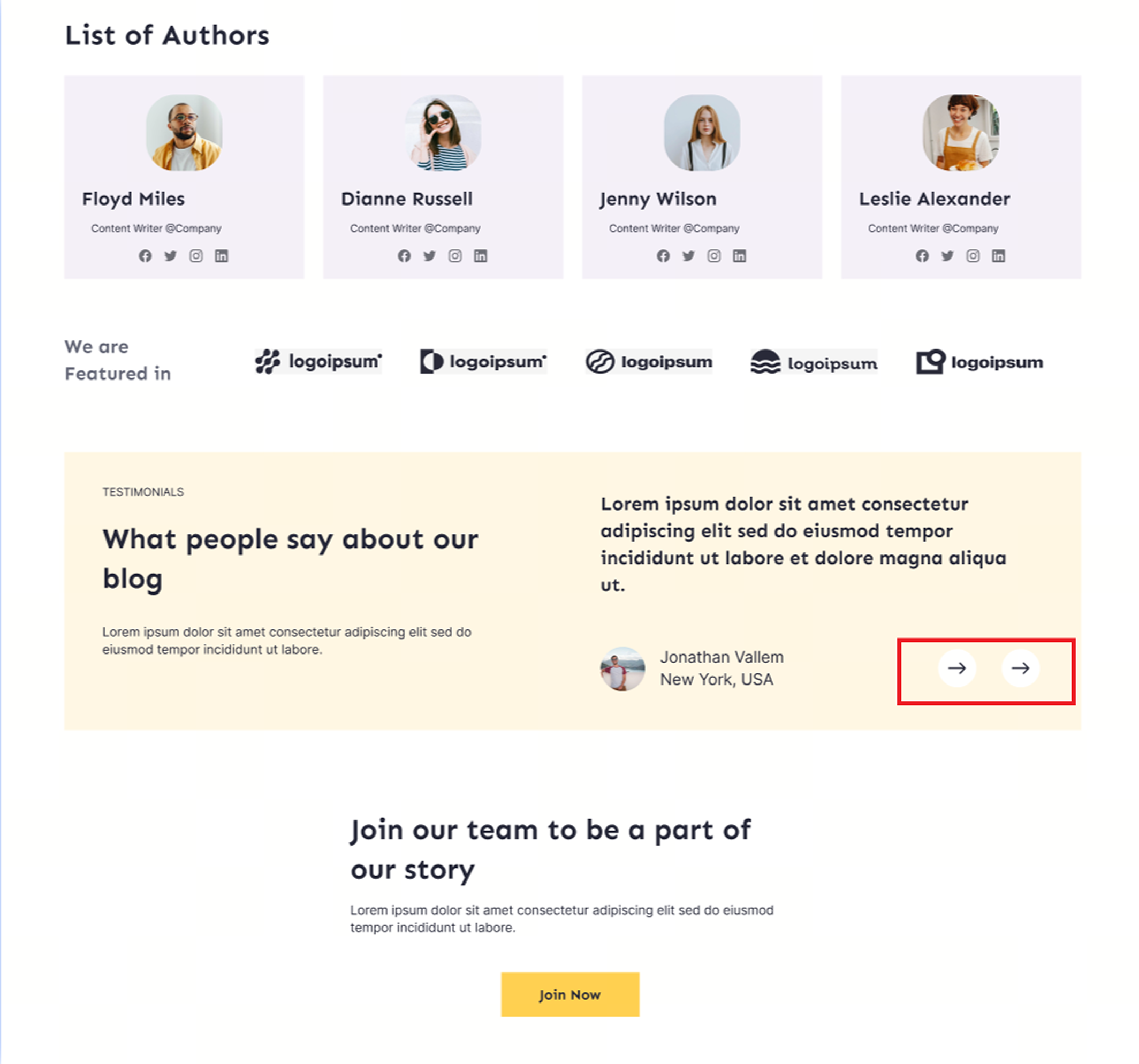}{Identical arrows for opposite actions}{GPT-6 Astra}
\end{center}

\subsection{Placeholder-looking icons}
Solid squares appear beside the gender and location options and in
the password field, where the icons would normally communicate
selection, location, or password visibility.

\begin{center}
  \failurecrop{0.46}{168bp}{7.28bp 177.06bp 0bp 222.06bp}
    {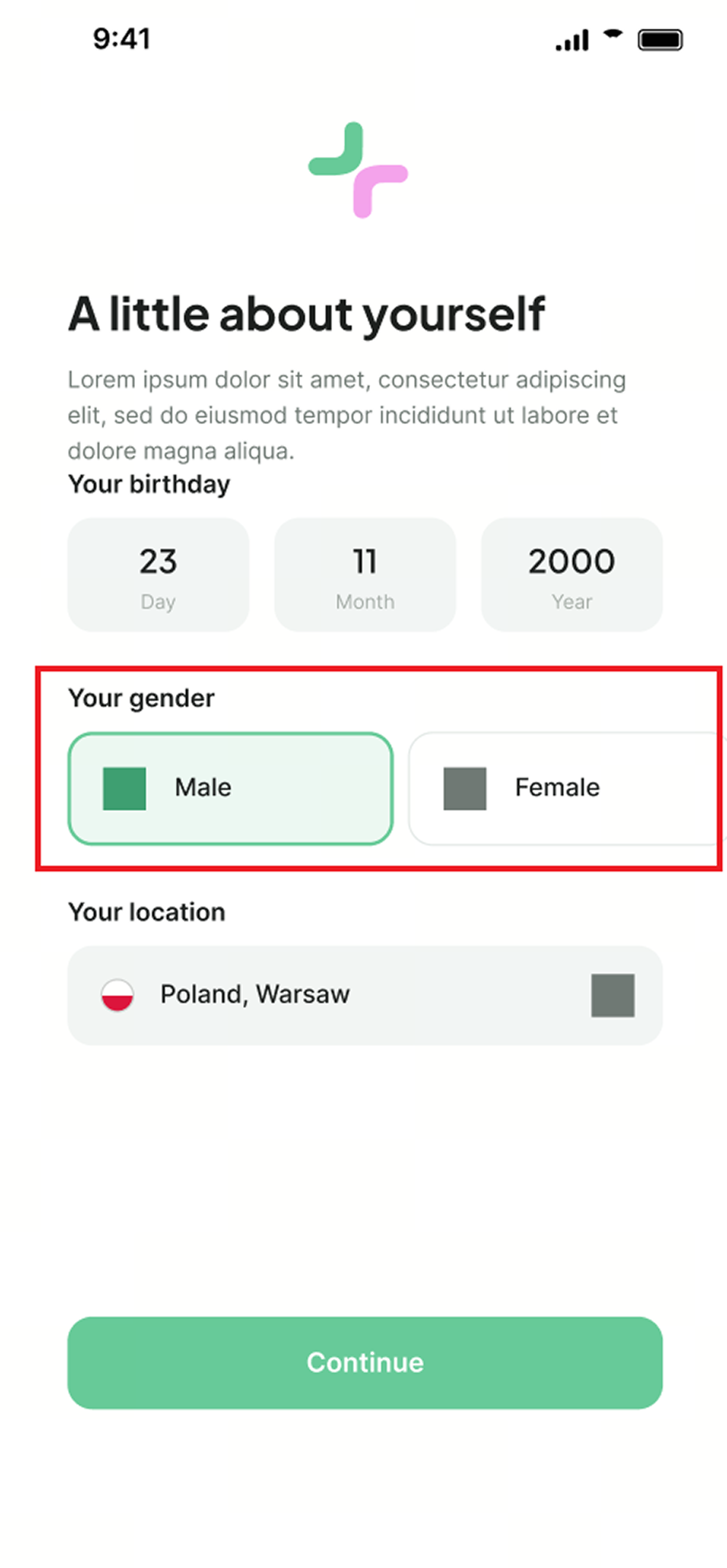}{Solid squares beside options}{Qwen 3.8 Max}\hspace{0.04\linewidth}%
  \failurecrop{0.46}{168bp}{10.05bp 266.43bp 8.62bp 98.84bp}
    {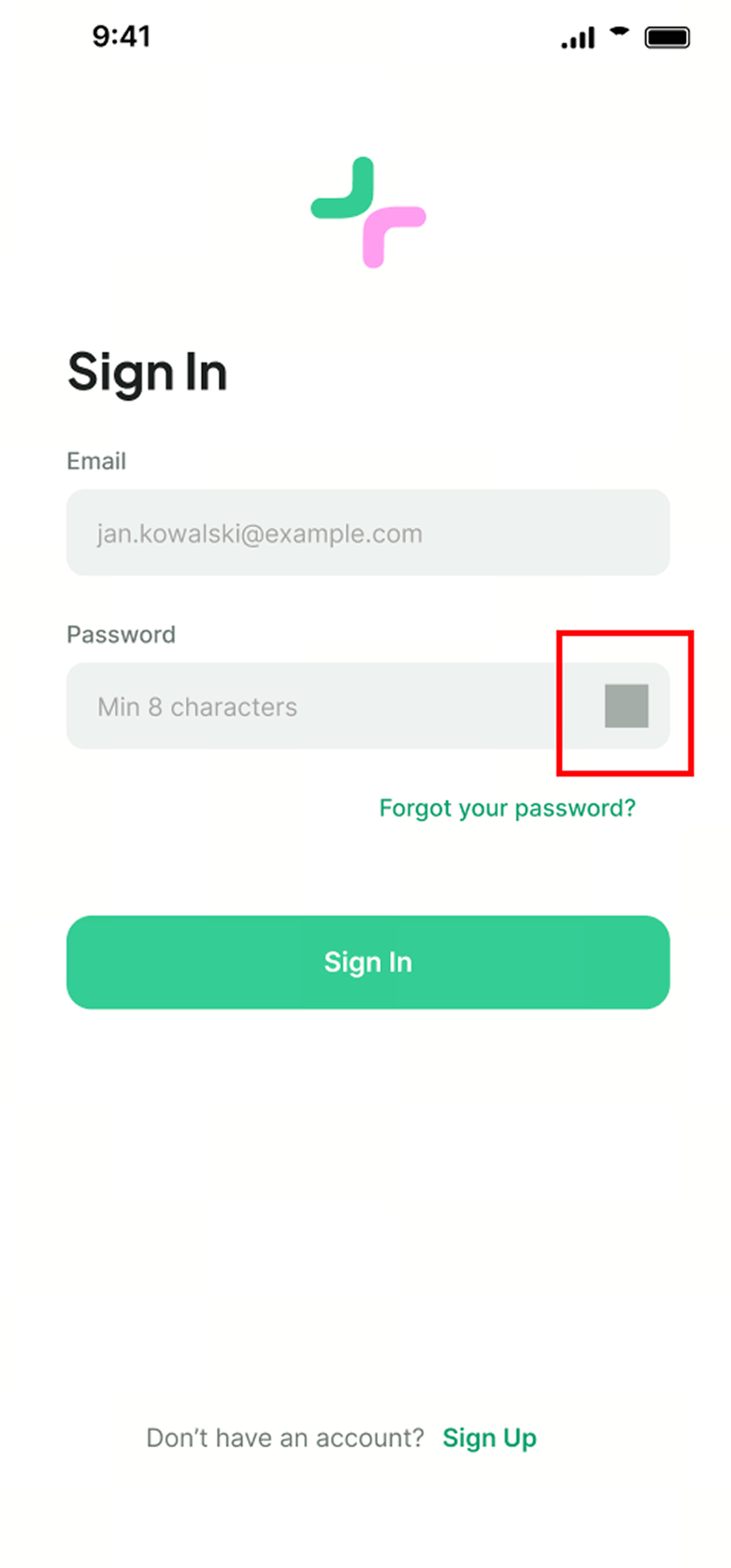}{Solid square in the password field}{Qwen 3.8 Max}
\end{center}

\clearpage

\subsection{No current-state indicator}
The rewards card reports progress of 4/8, but all eight cup icons look
identical, so the icon sequence does not show which rewards have been
earned. The Trending screen illustrates a weaker form of the problem:
the current tab is indicated only by a brighter label.

\begin{center}
  \failurecrop{0.46}{164bp}{3.20bp 376.32bp 1.60bp 35.08bp}
    {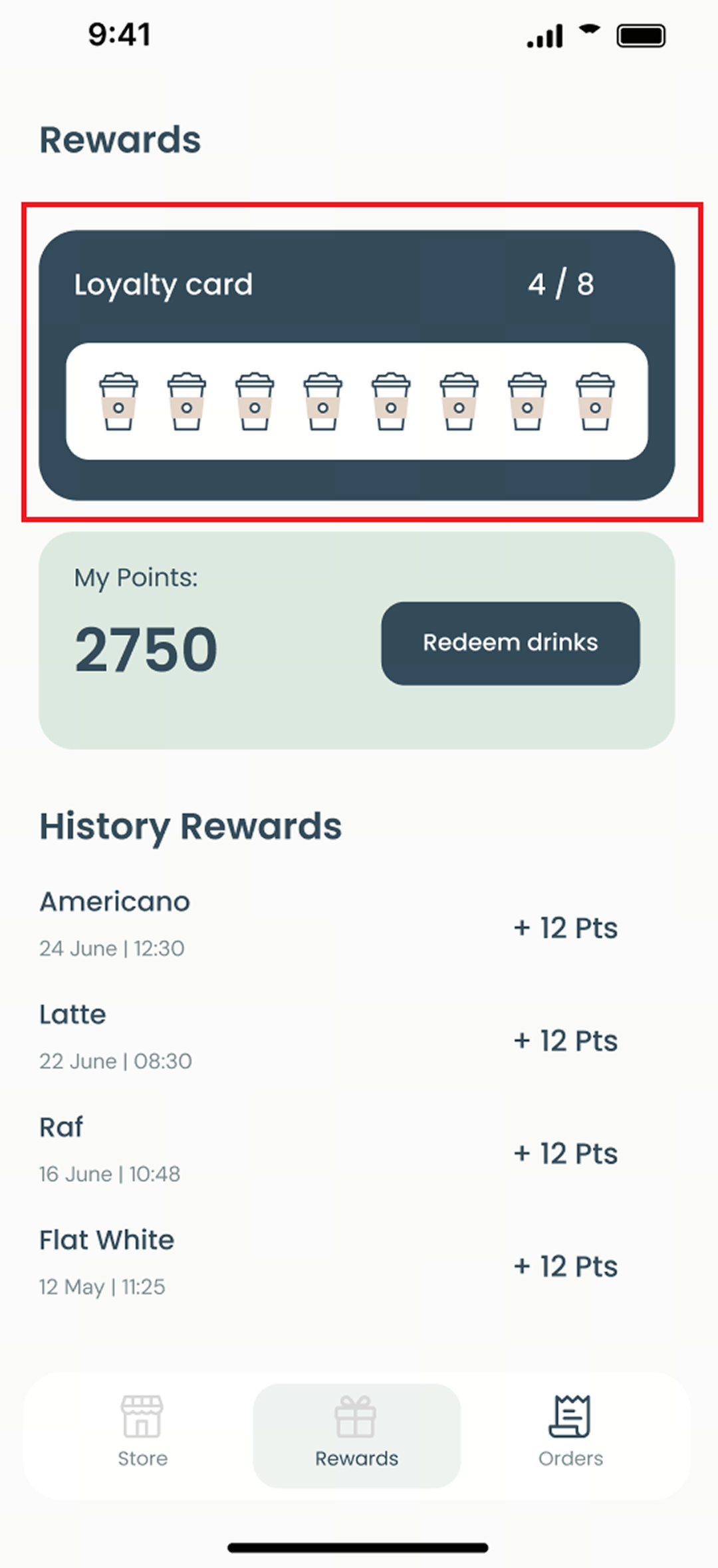}{Reward icons do not distinguish earned progress}{GPT-6 Astra}\hspace{0.04\linewidth}%
  \failurecrop{0.46}{164bp}{0bp 0bp 0bp 476.08bp}
    {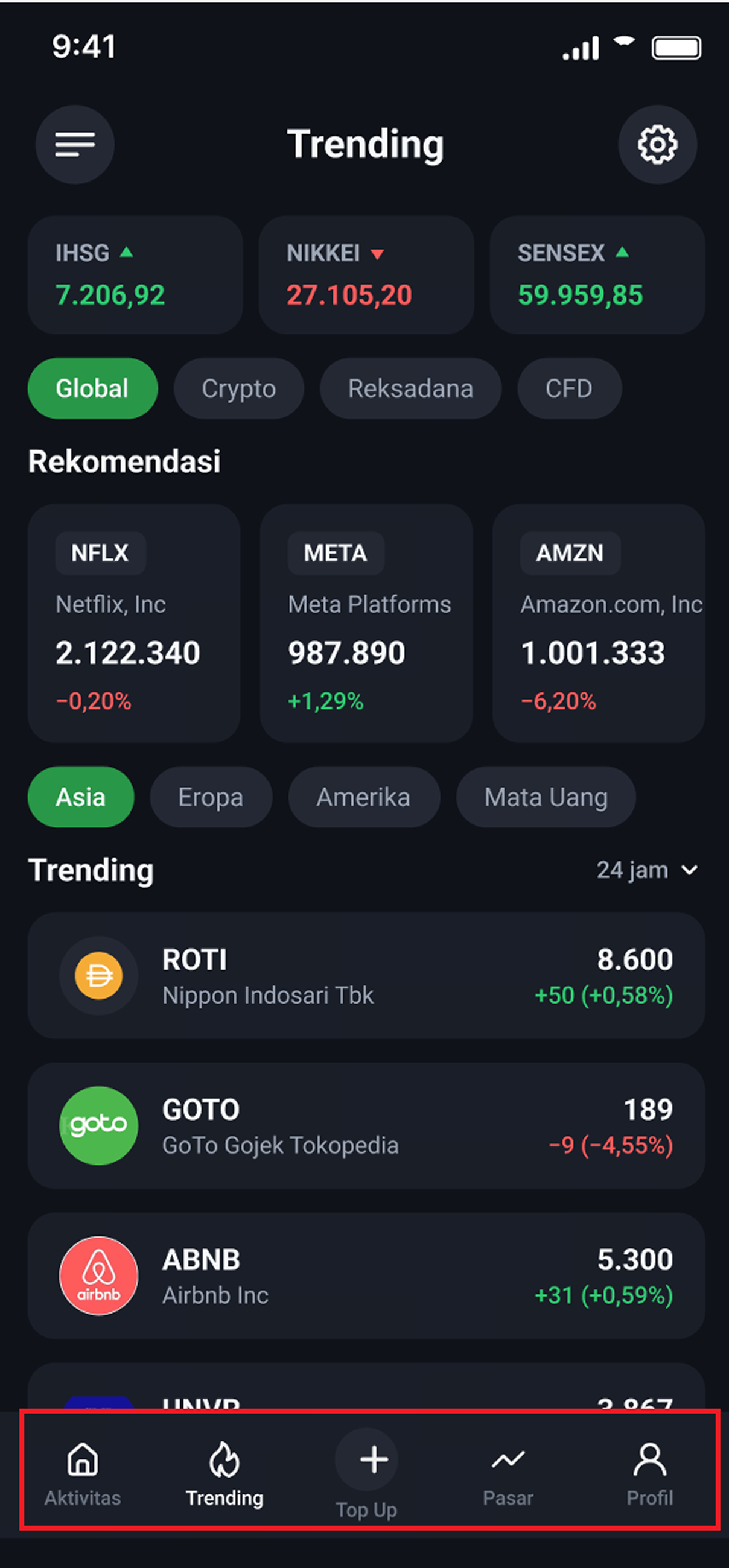}{Subtle active-tab label}{GPT-6.1 Sol}
\end{center}

\subsection{Low text contrast}
The white onboarding subtitle sits on a light-gray background,
making the text difficult to read. The product grid shows a related
problem with visual differentiation: \emph{Unavailable} and
\emph{Bestseller} use the same muted tag treatment even though they
communicate different information.

\begin{center}
  \failurecrop{0.46}{182bp}{4.65bp 145.32bp 4.65bp 267.45bp}
    {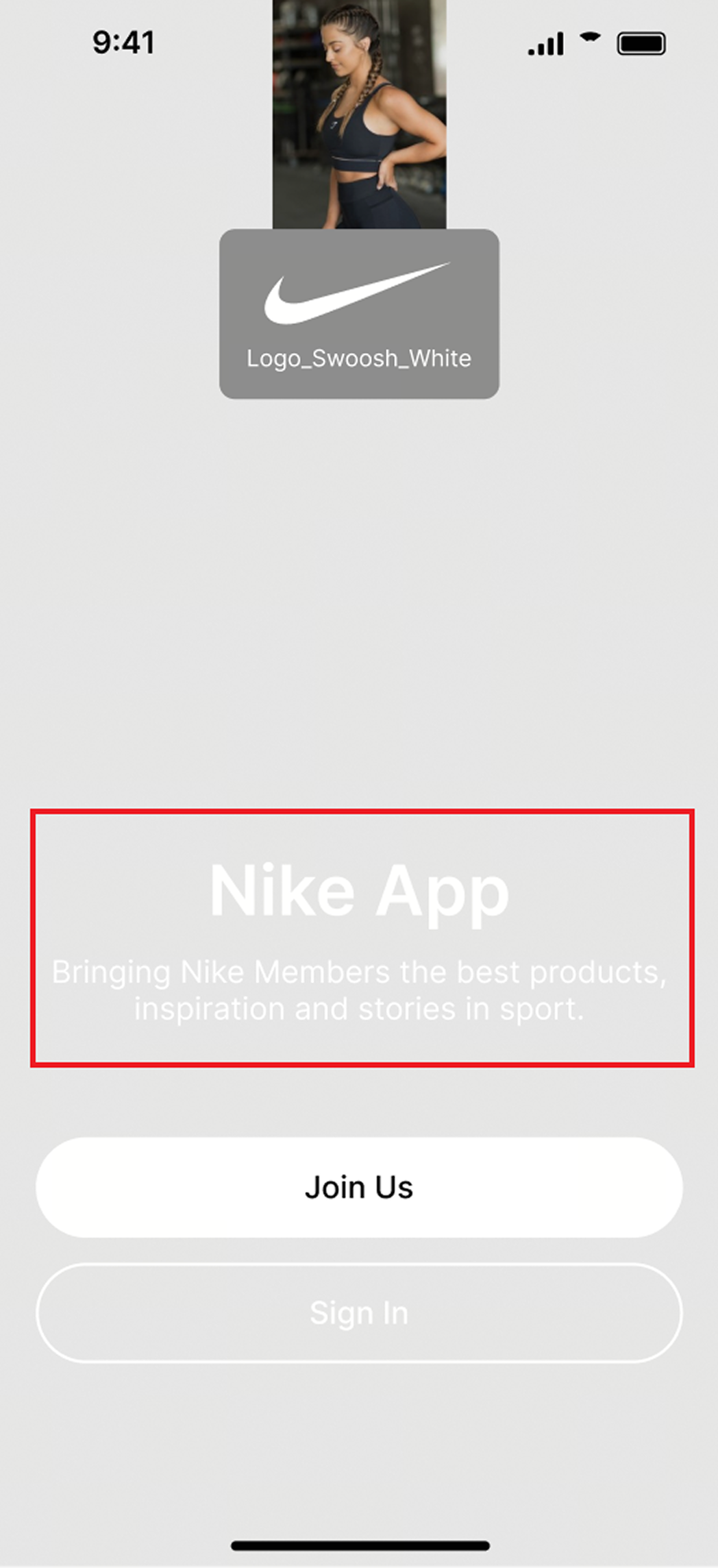}{White subtitle on a light-gray background}{Gemini 3.5 Flash}\hspace{0.04\linewidth}%
  \failurecrop{0.46}{182bp}{3.51bp 323.24bp 15.80bp 86.08bp}
    {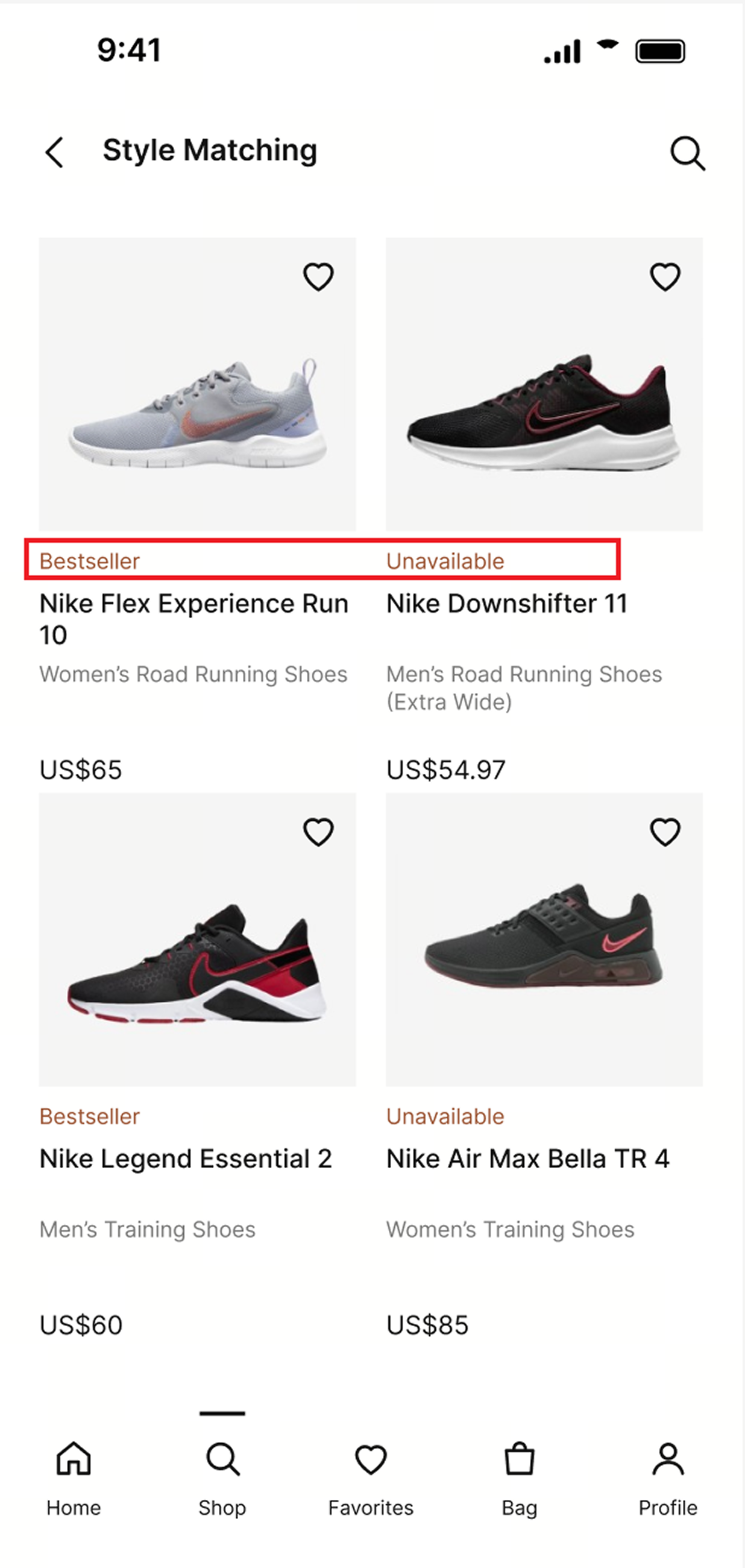}{Shared styling for availability and bestseller tags}{GPT-6 Astra}
\end{center}

\clearpage

\subsection{Domain knowledge unclear}
The generated stock chart shows red and green candle bodies without
visible wicks, so the high and low of each interval cannot be read
separately from its open and close. The reference chart shows how
candlestick charts encode those values: the body spans the open and
close, and the wick endpoints mark the high and low.

\begin{center}
  \failurecrop{0.47}{224bp}{0bp 78.76bp 0bp 153.70bp}
    {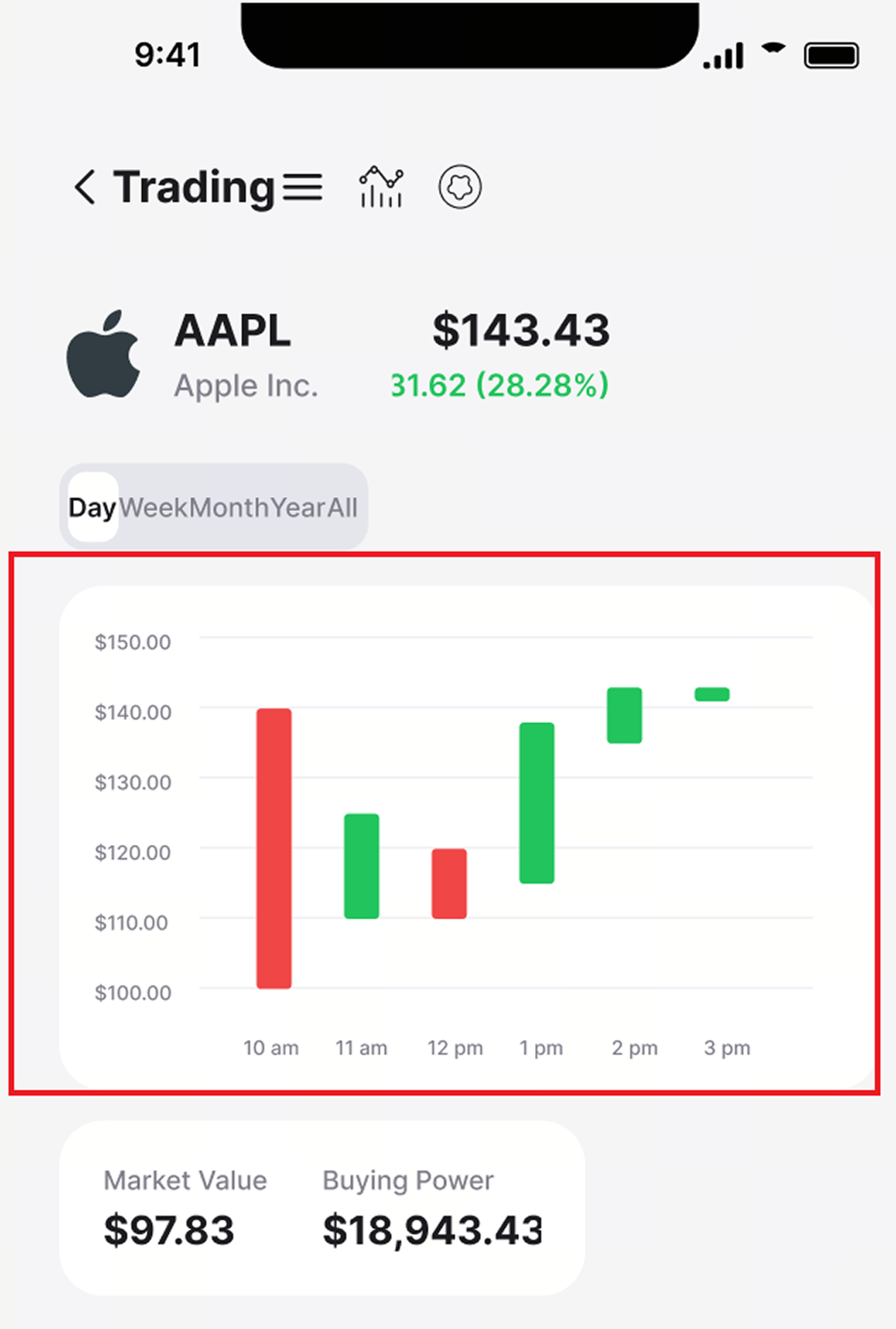}{Stock chart with candle bodies and no visible wicks}{Gemini 3.5 Flash}\hspace{0.035\linewidth}%
  \failurecrop{0.47}{224bp}{0bp 90bp 0bp 0bp}
    {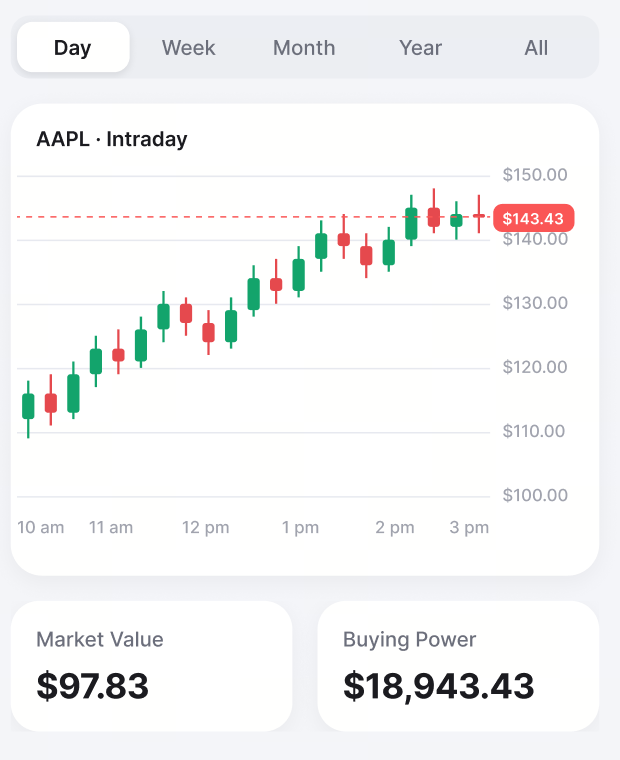}{Candlestick chart with visible high and low wicks}{Claude Code (Opus 5.5)}
\end{center}
\endgroup

\end{document}